\documentclass[letterpaper,12pt]{article}

\usepackage{times}
\usepackage[margin=1in]{geometry}

\usepackage{graphicx}
\DeclareGraphicsExtensions{.pdf,.jpeg,.png}
\usepackage{cite}
\usepackage{amsmath}
\usepackage{amssymb}
\usepackage{amsfonts}
\usepackage{titlesec}
\usepackage{enumerate}
\newtheorem{theorem}{Theorem}
\usepackage[linesnumbered,ruled]{algorithm2e}
\usepackage[justification=centering]{caption}
\titlelabel{\thetitle.\quad}
\titleformat*{\section}{\normalfont\bfseries}
\titleformat*{\subsection}{\normalfont\bfseries}
\titleformat*{\subsubsection}{\normalfont\bfseries}
\titleformat*{\paragraph}{\normalfont\bfseries}
\titleformat*{\subparagraph}{\normalfont\bfseries}
\newtheorem{definition}{Definition}
\newtheorem{remark}{Remark}
\newtheorem{proof}{Proof}
\usepackage{hyperref}
\hypersetup{hypertex=true,
	colorlinks=true,
	linkcolor=red,
	anchorcolor=blue,
	citecolor=green}
\usepackage[all]{xy}
\usepackage{arydshln}

\begin{document}
\date{}

\title{The Geometry and Kinematics of the Matrix Lie Group $SE_K(3)$}

\author{Yarong Luo, 
	yarongluo@whu.edu.cn\\
	Mengyuan Wang,
	mengyuan\_wang@whu.edu.cn\\
 Chi Guo,
 guochi@whu.edu.cn\\
 GNSS Research Center, Wuhan University
}



\maketitle

\thispagestyle{empty}

\noindent
{\bf\normalsize Abstract}\newline
{Currently state estimation is very important for the robotics, and the uncertainty representation based Lie group is natural for the state estimation problem. It is necessary to exploit the geometry and kinematic of matrix Lie group sufficiently. Therefore, this note gives a detailed derivation of the recently proposed matrix Lie group $SE_K(3)$ for the first time, our results extend the results in Barfoot~\cite{barfoot2017state}. Then the closed form for the inverse Jacobian in $SE_K(3)$ is derived. Next, we give detailed derivation of the composition of general poses. We also describe the situations where this group is suitable for state representation. Finally, we have developed code based on Matlab framework for quickly implementing and testing. 
} \vspace{2ex}

\noindent
{\bf\normalsize Key Words}\newline
{matrix Lie group $SE_K(3)$, Lie algebra $\mathfrak{se}_k(3)$,  geometry,   kinematics, state estimation}

\section{Introduction}
It is common to estimate the attitude, velocity and position for inertial navigation in mobile robotics. However, there is no much theory about the description of the extended poses. Alex Barrau proposed the matrix Lie group $SE_K(d)(K\in\mathbb{N},d\in \mathbb{N_+})$ which is called group of $K$ direct isometries to describe extended poses \cite{barrau2015non,barrau2015ekf} in the invariant extended Kalman filter algorithm\cite{barrau2016invariant}. This group seems to include all the Lie groups currently applied in the state estimation problem for robotics. The group $SE_K(d)$ allows recovering $SE_K(3)$ with $d=3$, and the theory of $SE_K(3)$ is further developed recently in contact-aided invariant extended Kalman filter for the legged robot state estimation\cite{Ross2020Contact}. The multiple direct spatial isometries group $SE_K(2)$ was also proposed for two dimensional state estimation problem in \cite{barrau2015ekf}. 
When $K$ equals to 0, the group $SE_K(3)$ reduced to the Matrix Lie group $SO(3)$ (the Special Orthogonal group) which represents the rotation matrices, 
and when $K$ equals to 1, it reduced to the Matrix Lie group $SE(3)$ (the Special Euclidean group) which is usually used to represent the pose (orientation, position) in state estimation. More details about these two special matrix Lie groups can be found in \cite{barfoot2017state}. 
When $K$ equals to 2, the group $SE_K(3)$ is called group of double direct isometries\cite{barrau2015non} and denoted as $SE_2(3)$. The matrix Lie group  $SE_2(3)$ has been used in many applications, such as solving the  localization problem from noisy inertial sensors and a noisy GPS by using $SE_2(3)$\cite{barrau2014invariant}, representing the extended poses (orientation, position,velocity) in 3D inertial navigation\cite{barrau2019linear} and  leveraging the theory of preintegration on manifolds that account for rotating earth\cite{bonnabel2020mathematical}. In most invariant filtering theory works, $SE_k(3)$ is proposed to resolve the consistency issues of KF-like problem\cite{barrau2015ekf}\cite{brossard2018unscented}.


The contribution in this note including:
\begin{itemize}
	\item the geometry and kinematics of the matrix Lie group $SE_K(3)$ are given similar to the matrix Lie group $SO(3)$ and $SE(3)$.
	\item the composition of general poses is derived in detail.
	\item the monotonicity of the uncertainty propagation on matrix Lie group $SE_K(3)$ is investigated. 
	\item the applications about the $SE_K(3)$ group are also discussed. 
	\item the simulation code about the manipulations of this group is released at \url{https://github.com/LYRen1900/Matrix-Lie-group-and-its-application}.
\end{itemize}




\section{Geometry of Matrix Lie Group $SE_K(3)$}
This section gives detailed derivation about the geometry of the $K$ direct isometries group $SE_K(3)$ by analogy with matrix Lie group $SE(3)$ and $SO(3)$ which have been shown in Barfoot's book~\cite{barfoot2017state}.
\subsection{Group of K Direct Isometries $SE_K(3)$}
The group represents the space of matrices that apply a rigid body rotation and K translations to points in $\mathbb{R}^3$. The group $SO(3)$ and the three dimensional vector space $\mathbb{R}^3$ is the subgroup of $SE_K(3)$.
Moreover, the group $SE_K(3)$ has the structure of the semidirect product of $SO(3)$ by $\underbrace{\mathbb{R}^3\times \mathbb{R}^3 \times \cdots \times \mathbb{R}^3}_K$ and it can be expressed as:
\begin{equation}\label{semedirect}
	SE_K(3)=SO(3) \propto  \underbrace{\mathbb{R}^3\times \mathbb{R}^3 \times \cdots \times \mathbb{R}^3}_K
\end{equation}
 The geometric meaning of the above semi-direct product is the rotation acting on the $K$ translations.
 
Formally, the matrix Lie group $SE_K(3)$ is defined as follows:
\begin{equation}\label{group_defition}
SE_K(3):=\left\{T=\begin{bmatrix}
R&p_1&p_2& \cdots & p_K\\
\bf{0}^T&1&0&\cdots &0\\
\bf{0}^T&0&1&\cdots&0\\
\vdots&\vdots&\vdots&\ddots&\vdots\\
\bf{0}^T&0&0&\cdots&1
\end{bmatrix}\in \mathbb{R}^{(K+3)\times (K+3)}|R\in SO(3),p_k\in \mathbb{R}^3,k=1,\cdots,K\right\}
\end{equation}
where SO(3) is the space of valid rotation matrices:
\begin{equation}\label{so_3_definition}
SO(3):=\left\{ R\in \mathbb{R}^{3\times 3}|RR^T=I_3,\det R=1\right\}
\end{equation}
and $I_d$ is the identity matrix of dimension d.

The identity element of the group is the $(K+3)\times (K+3)$ identity matrix. The matrix multiplication provides the group composition operation of any two elements in Lie group $SE_K(3)$. The closure for $SE_K(3)$ can be verified by multiplication,
\begin{equation}\label{closure_multiplycation}
\begin{aligned}
T_1T_2&=\begin{bmatrix}
R_1&{p_1}_1&{p_1}_2& \cdots & {p_1}_K\\
\bf{0}^T&1&0&\cdots &0\\
\bf{0}^T&0&1&\cdots&0\\
\vdots&\vdots&\vdots&\ddots&\vdots\\
\bf{0}^T&0&0&\cdot&1
\end{bmatrix}\begin{bmatrix}
R_2&{p_2}_1&{p_2}_2& \cdots & {p_2}_K\\
\bf{0}^T&1&0&\cdots &0\\
\bf{0}^T&0&1&\cdots&0\\
\vdots&\vdots&\vdots&\ddots&\vdots\\
\bf{0}^T&0&0&\cdots&1
\end{bmatrix}\\
&=\begin{bmatrix}
R_1R_2&R_1{p_2}_1+{p_1}_1&R_1{p_2}_2+{p_1}_2& \cdots & R_1{p_2}_K+{p_1}_K\\
\bf{0}^T&1&0&\cdots &0\\
\bf{0}^T&0&1&\cdots&0\\
\vdots&\vdots&\vdots&\ddots&\vdots\\
\bf{0}^T&0&0&\cdots&1
\end{bmatrix}\in SE_K(3)
\end{aligned}
\end{equation}
as a result of the closure $R_1R_2\in SO(3)$ and $R_1{p_2}_k+{p_1}_k\in \mathbb{R}^3,k=1,\cdots,K$.

For invertibility, the inverse of the element is given as 
\begin{equation}\label{inverse_of_group}
T^{-1}=\begin{bmatrix}
R^{-1}&-R^{-1}p_1&-R^{-1}p_2& \cdots & -R^{-1}p_K\\
\bf{0}^T&1&0&\cdots &0\\
\bf{0}^T&0&1&\cdots&0\\
\vdots&\vdots&\vdots&\ddots&\vdots\\
\bf{0}^T&0&0&\cdots&1
\end{bmatrix}\in SE_K(3)
\end{equation}
\subsection{Lie algebra $\mathfrak{se}_k(3)$}
The tangent space of the matrix Lie group $SE_K(3)$ at the identity element is the Lie algebra $\mathfrak{se}_3(k)$, which is also a vectorspace that completely captures the local structure of the group.
The Lie algebra associated with $SE_K(3)$ is given by
\begin{equation*}
\begin{aligned}
\text{vectorspace}:&\quad \mathfrak{se}_k(3)=\left\{S=\mathcal{L}(\xi)\in \mathbb{R}^{(K+3)\times (K+3)}|\xi\in\mathbb{R}^{3(K+1)}\right\}\\
\text{field}:&\quad \mathbb{R}\\
\text{Lie bracket}:&\quad [S_1,S_2]=S_1S_2-S_2S_1
\end{aligned}
\end{equation*}
This space of matrices is isomorphic to $\mathbb{R}^{3(K+1)}$, and aims at to identify $\mathfrak{se}_k(3)$ to $\mathbb{R}^{3(K+1)}$. We define a linear map $\mathcal{L}:\mathbb{R}^{3(K+1)}\rightarrow \mathbb{R}^{(K+3)\times (K+3)}$ that converts the Euclidean vector to the matrix forms:
\begin{equation}\label{lie_algebra_se_3_k}
\mathcal{L}(\xi)=\mathcal{L}\left(\begin{bmatrix}
\phi\\t_1\\t_2\\ \vdots \\t_K
\end{bmatrix}\right)=\begin{bmatrix}
\phi_{\wedge}&t_1&t_2& \cdots & t_K\\
\bf{0}^T&0&0&\cdots &0\\
\bf{0}^T&0&0&\cdots&0\\
\vdots&\vdots&\vdots&\ddots&\vdots\\
\bf{0}^T&0&0&\cdots&0
\end{bmatrix},\phi,t_1,t_2,\cdots,t_K\in\mathbb{R}^3
\end{equation}
where $\phi_{\wedge}$ denotes the skew-symmetric matrix of a vector $\phi=\begin{pmatrix}
\phi_1 &\phi_2&\phi_3
\end{pmatrix}^T\in\mathbb{R}^3$.

Let $S_1=\mathcal{L}(\xi_1),S_2=\mathcal{L}(\xi_2)\in \mathfrak{se}_k(3)$, then, for the closure property of \text{the Lie bracket}, we can get
\begin{equation}\label{closure_property}
[S_1,S_2]=S_1S_2-S_2S_1=\mathcal{L}(\xi_1)\mathcal{L}(\xi_2)-\mathcal{L}(\xi_2)\mathcal{L}(\xi_1)=\mathcal{L}\left(\mathfrak{L}(\xi_1)\xi_2\right)\in \mathfrak{se}_k(3)
\end{equation}
where $\mathfrak{L}$ is also defined as a linear map $\mathfrak{L}:\mathbb{R}^{3(K+1)}\rightarrow \mathbb{R}^{(K+3)\times (K+3)}$ that converts the Euclidean vector to the matrix forms:
\begin{equation}\label{adjoint_represnetation_algebra}
\mathfrak{L}(\xi)=\mathfrak{L}\left(\begin{bmatrix}
\phi\\t_1\\t_2\\ \vdots \\t_K
\end{bmatrix}\right)=\begin{bmatrix}
\phi_{\wedge}&0&0&\cdots&0\\
(t_1)_{\wedge}&\phi_{\wedge}&0&\cdots&0\\
(t_2)_{\wedge}&0&\phi_{\wedge}&\cdots&0\\
\vdots&\vdots&\vdots&\ddots&\vdots\\
(t_K)_{\wedge}&0&0&\cdots&\phi_{\wedge}
\end{bmatrix}\in\mathbb{R}^{3(K+1)\times3(K+1)}
\end{equation}

The Lie algebra $\mathfrak{se}_k(3)$ is the set of $(K+3)\times (K+3)$ matrices corresponding to differential translations and rotations. Therefore, there are thus $3(K+1)$ generators of this Lie algebra:
\begin{equation}\label{generators}
\begin{aligned}
G_1&=\begin{bmatrix}
0&0&0&0_{1\times 3K}\\
0&0&-1&0_{1\times 3K}\\
0&1&0&0_{1\times 3K}\\
0_{3K\times 1}&0_{3K\times 1}&0_{3K\times 1}&0_{3K\times 3K}
\end{bmatrix},G_2=\begin{bmatrix}
0&0&1&0_{1\times 3K}\\
0&0&0&0_{1\times 3K}\\
-1&&0&0_{1\times 3K}\\
0_{3K\times 1}&0_{3K\times 1}&0_{3K\times 1}&0_{3K\times 3K}
\end{bmatrix}\\
G_3&=\begin{bmatrix}
0&-1&0&0_{1\times 3K}\\
1&0&0&0_{1\times 3K}\\
0&0&0&0_{1\times 3K}\\
0_{3K\times 1}&0_{3K\times 1}&0_{3K\times 1}&0_{3K\times 3K}
\end{bmatrix},
G_{3k+1}=\begin{bmatrix}
0_{3\times 3}& \begin{array}{ccc}
0_{1\times (k-1)} & 1& 0_{1\times {(K-k)}}\\
0_{1\times (k-1)} & 0& 0_{1\times {(K-k)}}\\
0_{1\times (k-1)} & 0& 0_{1\times {(K-k)}}
\end{array}\\
0_{3K\times 3}&0_{3K\times 3K}
\end{bmatrix}\\
G_{3k+2}&=\begin{bmatrix}
0_{3\times 3}& \begin{array}{ccc}
0_{1\times (k-1)} & 0& 0_{1\times {(K-k)}}\\
0_{1\times (k-1)} & 1& 0_{1\times {(K-k)}}\\
0_{1\times (k-1)} & 0& 0_{1\times {(K-k)}}
\end{array}\\
0_{3K\times 3}&0_{3K\times 3K}
\end{bmatrix},G_{3k+3}=\begin{bmatrix}
0_{3\times 3}& \begin{array}{ccc}
0_{1\times (k-1)} & 0& 0_{1\times {(K-k)}}\\
0_{1\times (k-1)} & 0& 0_{1\times {(K-k)}}\\
0_{1\times (k-1)} & 1& 0_{1\times {(K-k)}}
\end{array}\\
0_{3K\times 3}&0_{3K\times 3K}
\end{bmatrix}\\
&k=1,\cdots,K
\end{aligned}
\end{equation}

Any element of the Lie algebra $\mathfrak{se}_k(3)$ can be represented as a linear combination of the generators:
\begin{equation}\label{linear_combination}
\begin{aligned}
\begin{bmatrix}
\phi^T &t_1^T&t_2^T& \cdots & t_K^T
\end{bmatrix}^T &\in \mathbb{R}^{3(K+1)}\\
\phi_1G_1+\phi_2G_2+\phi_3G_3+\sum_{k=1}^{K}\left({t_k}_1G_{3k+1}+{t_k}_2G_{3k+2}+{t_k}_3G_{3k+3}\right) &\in \mathfrak{se}_k(3)
\end{aligned}
\end{equation}

Consequently, the Lie algebra space is a $3(K+1)$-dimensional vector space with basis elements $\{G_1,\cdots,G_{3(K+1)}\}$, and there is a linear isomorphism between $\mathfrak{se}_k(3)$ and $\mathbb{R}^{3(K+1)}$ that we denote as follows: $\mathcal{L}:\mathbb{R}^{3(K+1)}\rightarrow \mathbb{R}^{(K+3)\times (K+3)}$ and $\mathcal{L}^{-1}:\mathbb{R}^{(K+3)\times (K+3)} \rightarrow \mathbb{R}^{3(K+1)}$. The isomorphism property will allow us to express the differential calculus of the matrix Lie group computation in vector form with the minimal number of parameters $3(K+1)$, as opposed to considering the matrices with $(K+3)^2$ coefficients. 
Moreover, the Euclidean space $\mathbb{R}^{3(K+1)}$ is sometimes called the Lie algebra of the Lie group $SE_K(3)$ in some literatures.
\subsection{Exponential Map}
The exponential map from the Lie algebra $\mathfrak{se}_k(3)$ to the Lie group $SE_K(3)$ is the simple extension of the Lie algebra $\mathfrak{se}(3)$. Consequently, given the vector $\xi\in\mathbb{R}^{3(K+1)}$ and the Lie algebra  $\mathcal{L}(\xi)$, the Lie group can be given by 
\begin{equation}\label{exponential_map_sek_3_to_gourp}
\begin{aligned}
T&=Exp(\xi)=\exp_G(\mathcal{L}(\xi))=\begin{bmatrix}
\exp_G(\phi_{\wedge})&J_lt_1&J_lt_2& \cdots & J_lt_K\\
\bf{0}^T&1&0&\cdots &0\\
\bf{0}^T&0&1&\cdots&0\\
\vdots&\vdots&\vdots&\ddots&\vdots\\
\bf{0}^T&0&0&\cdots&1
\end{bmatrix}\\
&=\begin{bmatrix}
R&p_1&p_2& \cdots & p_K\\
\bf{0}^T&1&0&\cdots &0\\
\bf{0}^T&0&1&\cdots&0\\
\vdots&\vdots&\vdots&\ddots&\vdots\\
\bf{0}^T&0&0&\cdots&1
\end{bmatrix}\in \mathbb{R}^{(K+3)\times (K+3)},R\in SO(3),p_k\in \mathbb{R}^3,k=1,\cdots,K
\end{aligned}
\end{equation}
where $Exp$ denotes the map that maps the vector $\xi$ to $SE_K(3)$,  $\exp_G$ denotes the matrix exponential, 
\begin{equation}\label{R_SO3_exponential}
\begin{aligned}
R=R(\phi)=\exp_G(\phi_{\wedge})&=I_3+\sin\theta u_{\wedge}+(1-\cos\theta)u_{\wedge}^2\\
&=I_3+\sin\theta u_{\wedge}+(1-\cos\theta)(-I_3+uu^T)\\
&=\cos\theta I_3+(1-\cos\theta)uu^T+\sin\theta u_{\wedge}\\
&=I_3+\frac{\sin\theta}{\theta}\phi_{\wedge}+\frac{1-\cos\theta}{\theta^2}\phi_{\wedge}^2,\theta=||\phi||,u=\phi/\theta\\
\end{aligned}
\end{equation}
\begin{equation}\label{unit_vector_high_order}
u_{\wedge}^3=-u_{\wedge}
\end{equation}
\begin{equation}\label{left_Jacobian_vector}
p_k=J_lt_k\in \mathbb{R}^3,k=1,\cdots,K
\end{equation}
and $J_l$ is the left Jacobian matrix of the rotation matrices group $SO(3)$ which is given by 
\begin{equation}\label{left_Jacobian}
\begin{aligned}
J_l=J_l(\phi)&=\sum_{n=0}^{\infty}\frac{1}{(n+1)!}(\phi_{\wedge})^n=\int_{0}^{1}R^{\alpha}d\alpha\\
&=I_3+\frac{1-\cos\theta}{\theta}u_{\wedge}+(1-\frac{\sin\theta}{\theta})u_{\wedge}^2\\
&=I_3+\frac{1-\cos\theta}{\theta}u_{\wedge}+(1-\frac{\sin\theta}{\theta})(-I_3+uu^T)\\
&=\frac{\sin\theta}{\theta}I_3+(1-\frac{\sin\theta}{\theta})uu^T+\frac{1-\cos\theta}{\theta}u_{\wedge}\\
&=I_3+\frac{1-\cos\theta}{\theta^2}\phi_{\wedge}+\frac{\theta-\sin\theta}{\theta^3}\phi_{\wedge}^2
\end{aligned}
\end{equation}
and the inverse of the left Jacobian matrix $J_l$ is 
\begin{equation}\label{inverse_left_Jacobian_matrix}
\begin{aligned}
J_l^{-1}=J_l^{-1}(\phi)&=\sum_{n=0}^{\infty}\frac{B_n}{n!}(\phi_{\wedge})^n=I_3-\frac{\theta}{2}u_{\wedge}+(1-\frac{\theta}{2}\cot\frac{\theta}{2})u_{\wedge}u_{\wedge}\\
&=I_3-\frac{\theta}{2}u_{\wedge}+(1-\frac{\theta}{2}\cot\frac{\theta}{2})(-I_3+uu^T)\\
&=\frac{\theta}{2}\cot\frac{\theta}{2}I_3+(1-\frac{\theta}{2}\cot\frac{\theta}{2})uu^T-\frac{\theta}{2}u_{\wedge}\\
&=I-\frac{1}{2}\phi_{\wedge}+\left(\frac{1}{\theta^2}-\frac{1+\cos\theta}{2\theta\sin\theta} \right)\phi_{\wedge}^2
\end{aligned}
\end{equation}
where $B_n (n=1,\cdots,\infty)$ are the Bernoulli numbers. 

According to the identity (\ref{unit_vector_high_order}), it is easy to verify that 
\begin{equation}\label{cayley_hamilton_rotation}
\theta^3(u_{\wedge}^3+u_{\wedge})=\phi_{\wedge}^3+\theta^2\phi_{\wedge}=0_{3\times 3}\Rightarrow \phi_{\wedge}^3=-\theta^2\phi_{\wedge}
\end{equation}

The identity relationship between the Rodriguez formula and the left Jacobian matrix can be proofed as following:
\begin{equation}\label{left_jacobian_rodriguez}
\begin{aligned}
J_l(R\phi)&=I_3+\frac{1-\cos\theta}{\theta^2}(R\phi)_{\wedge}+\frac{\theta-\sin\theta}{\theta^3}(R\phi)_{\wedge}^2\\
&=I_3+\frac{1-\cos\theta}{\theta^2}R\phi_{\wedge}R^T+\frac{\theta-\sin\theta}{\theta^3}R\phi_{\wedge}R^TR\phi_{\wedge}R^T\\
&=RR^T+\frac{1-\cos\theta}{\theta^2}R\phi_{\wedge}R^T+\frac{\theta-\sin\theta}{\theta^3}R\phi_{\wedge}^2R^T=RJ_l(\phi)R^T
\end{aligned}
\end{equation}
\begin{equation}\label{left_jacobian_rodriguez_2}
\begin{aligned}
&J_l(\phi)\phi_{\wedge}=\left(I_3+\frac{1-\cos\theta}{\theta^2}\phi_{\wedge}+\frac{\theta-\sin\theta}{\theta^3}\phi_{\wedge}^2\right)\phi_{\wedge}=\phi_{\wedge}+\frac{1-\cos\theta}{\theta^2}\phi_{\wedge}^2+\frac{\theta-\sin\theta}{\theta^3}\phi_{\wedge}^3\\
=&\phi_{\wedge}+\frac{1-\cos\theta}{\theta^2}\phi_{\wedge}^2+(\theta-\sin\theta)u_{\wedge}^3=\phi_{\wedge}+\frac{1-\cos\theta}{\theta^2}\phi_{\wedge}^2+(\theta-\sin\theta)(-u_{\wedge})\\
=&\frac{1-\cos\theta}{\theta^2}\phi_{\wedge}^2+\frac{\sin\theta}{\theta}\phi_{\wedge}=R(\phi)-I_3
\end{aligned}
\end{equation}

It is well known that
\begin{equation}\label{rotation_3d}
(R\phi)_{\wedge}=R(\phi_{\wedge})R^T,R\in SO(3), \phi\in\mathbb{R}^3
\end{equation}
Therefore, we can get
\begin{equation}\label{rotation_rotation}
\begin{aligned}
&R(R_1\phi)=I_3+\frac{\sin\theta}{\theta}(R_1\phi)_{\wedge}+\frac{1-\cos\theta}{\theta^2}(R_1\phi)_{\wedge}^2\\
=&R_1R_1^T+\frac{\sin\theta}{\theta}R_1(\phi_{\wedge})R_1^T+\frac{1-\cos\theta}{\theta^2}R_1(\phi_{\wedge}^2)R_1^T=R_1R(\phi)R_1^T, R_1\in SO(3),\phi\in \mathbb{R}^3
\end{aligned}
\end{equation}

Furthermore, we have that
\begin{equation}\label{rotation_ad}
\exp_G((R\phi)_{\wedge})=R\exp_G(\phi_{\wedge})R^T, R\in SO(3), \phi\in \mathbb{R}^3
\end{equation}
By equation (\ref{rotation_3d}), the proof is as follows:
\begin{equation}\label{rotation_ad_proof}
\begin{aligned}
&\exp_G((R\phi)_{\wedge})=\sum_{n=0}^{\infty}\frac{1}{n!}(R\phi)_{\wedge}^n=\sum_{n=0}^{\infty}\frac{1}{n!}(R(\phi_{\wedge})R^T)^n=\sum_{n=0}^{\infty}\frac{1}{n!}R(\phi_{\wedge})^nR^T\\
=&R\sum_{n=0}^{\infty}\frac{1}{n!}(\phi_{\wedge})^nR^T=R\exp_G(\phi_{\wedge})R^T
\end{aligned}
\end{equation}

we can go in the other direction but not uniquely use the logarithmic mapping:
\begin{equation}\label{logarithm_map_definition}
\xi=Log(T)=\mathcal{L}^{-1}(\log_G(T))
\end{equation}
where $Log$ denotes the map that maps the element $T$ of matrix Lie group $SE_K(3)$ to the vector $\xi$, and $\log_G$ denotes the matrix logarithm.

The exponential map from $\mathfrak{se}_k(3)$ to $SE_K(3)$ is surjective-only: every $T\in SE_K(3)$ can be generated by many $\xi\in \mathbb{R}^{3(K+1)}$.

We can also develop a direct series expression for T from the exponential map by using the useful identity 
\begin{equation}\label{direct_series_expression}
\mathcal{L}(\xi)^4+\theta^2\mathcal{L}(\xi)^2=S^4+\theta^2S^2=\bf{0}_{K+3}
\end{equation}
where $\xi=\begin{bmatrix}
\phi^T &t_1^T&t_2^T& \cdots & t_K^T
\end{bmatrix}^T \in \mathbb{R}^{3(K+1)}$. 
Expanding the series and using the identity to rewrite all quartic and higher terms in terms of lower-order terms, 
the closed form expression for the exponential map of $SE_K(3)$ is given by
\begin{equation}\label{Exponential_map_se_k_3}
\begin{aligned}
& T=Exp(\xi)=\exp_G(\mathcal{L}(\xi))=\exp_G(S)=\sum_{n=0}^{\infty}\frac{1}{n!}S^n\\
=&I_{K+3}+S+\frac{1}{2!}S^2+\frac{1}{3!}S^3+\frac{1}{4!}S^4+\frac{1}{5!}S^5+\cdots\\
=&I_{K+3}+S+\underbrace{\left(\frac{1}{2!}-\frac{1}{4!}\theta^2+\frac{1}{6!}\theta^4-\frac{1}{8!}\theta^6+\cdots \right)}_{\frac{1-\cos{\theta}}{\theta^2}} S^2\\
\quad
&+\underbrace{\left(\frac{1}{3!}-\frac{1}{5!}\theta^2+\frac{1}{7!}\theta^4-\frac{1}{9!}\theta^6+\cdots \right)}_{\frac{\theta-\sin{\theta}}{\theta^3}} S^3\\
=&I_{K+3}+S+\frac{1-\cos\theta}{\theta^2}S^2+\frac{\theta-\sin\theta}{\theta^3}S^3
\end{aligned}
\end{equation}
\subsection{Adjoints}
The Adjoint representation of a matrix Lie group $G$ on the Euclidean space $\mathbb{R}^{p}$ is defined as the linear operator $Ad_T$, it captures properties related to commutation:
\begin{equation}\label{adjoint_representation_definition}
\begin{aligned}
\forall T\in G,\xi\in \mathbb{R}^p  \quad &Exp(Ad_T\cdot \xi)=\exp_G(\mathcal{L}(Ad_T\cdot \xi))\\
:=&\exp_G(T\mathcal{L}(\xi)T^{-1})=\sum_{n=0}^{\infty}(T\mathcal{L}(\xi)T^{-1})^n\\
=&T\sum_{n=0}^{\infty}(\mathcal{L}(\xi))^nT^{-1}=T\exp_G(\mathcal{L}(\xi))T^{-1}=TExp(\xi)T^{-1}
\end{aligned}
\end{equation}
the Adjoint representation describes the effect of non-commutability matrix multiplication. 

Assuming we have a vector $\xi\in \mathbb{R}^{3(K+1)}$, the corresponded Lie algebra $\mathfrak{se}_k(3)$ and matrix Lie group $T\in SE_K(3)$, 
combining the nice property that $R\phi_{\wedge}R^{-1}=(R\phi)_{\wedge}$, the adjoin of an element of $SE_K(3)$ is given by analogy to $SE(3)$:
\begin{equation}\label{adjoint_representation_derivtive}
\begin{aligned}
&\qquad Exp(Ad_T\cdot \xi)=\exp_G(\mathcal{L}(Ad_T\cdot \xi))=\exp_G(T\mathcal{L}(\xi)T^{-1})\\
&=\exp_G\left(\begin{bmatrix}
R&p_1&p_2& \cdots & p_K\\
\bf{0}^T&1&0&\cdots &0\\
\bf{0}^T&0&1&\cdots&0\\
\vdots&\vdots&\vdots&\ddots&\vdots\\
\bf{0}^T&0&0&\cdot&1
\end{bmatrix}
\begin{bmatrix}
\phi_{\wedge}&t_1&t_2& \cdots & t_K\\
\bf{0}^T&0&0&\cdots &0\\
\bf{0}^T&0&0&\cdots&0\\
\vdots&\vdots&\vdots&\ddots&\vdots\\
\bf{0}^T&0&0&\cdot&0
\end{bmatrix}
\begin{bmatrix}
R^{-1}&-R^{-1}p_1&-R^{-1}p_2& \cdots & -R^{-1}p_K\\
\bf{0}^T&1&0&\cdots &0\\
\bf{0}^T&0&1&\cdots&0\\
\vdots&\vdots&\vdots&\ddots&\vdots\\
\bf{0}^T&0&0&\cdot&1
\end{bmatrix}
\right)\\
&=\exp_G\left(\begin{bmatrix}
R\phi_{\wedge}R^{-1}&-R\phi_{\wedge}R^{-1}p_1+Rt_1&-R\phi_{\wedge}R^{-1}p_2+Rt_2& \cdots & -R\phi_{\wedge}R^{-1}p_K+Rt_K\\
\bf{0}^T&0&0&\cdots &0\\
\bf{0}^T&0&0&\cdots&0\\
\vdots&\vdots&\vdots&\ddots&\vdots\\
\bf{0}^T&0&0&\cdots&0
\end{bmatrix}
\right)\\
&=\exp_G\left(\begin{bmatrix}
(R\phi)_{\wedge}&(p_1)_{\wedge}R\phi+Rt_1&(p_2)_{\wedge}R\phi+Rt_2& \cdots & (p_K)_{\wedge}R\phi+Rt_K\\
\bf{0}^T&0&0&\cdots &0\\
\bf{0}^T&0&0&\cdots&0\\
\vdots&\vdots&\vdots&\ddots&\vdots\\
\bf{0}^T&0&0&\cdots&0
\end{bmatrix}
\right)\\
&=\exp_G\left(\mathcal{L}\left(\begin{bmatrix}
R\phi\\
(p_1)_{\wedge}R\phi+Rt_1\\
(p_2)_{\wedge}R\phi+Rt_2\\
\vdots\\
(p_K)_{\wedge}R\phi+Rt_K
\end{bmatrix}\right)
\right)=Exp\left(\begin{bmatrix}
R&0&0&\cdots&0\\
(p_1)_{\wedge}R&R&0&\cdots&0\\
(p_2)_{\wedge}R&0&R&\cdots&0\\
\vdots&\vdots&\vdots&\ddots&\vdots\\
(p_K)_{\wedge}R&0&0&\cdots&R
\end{bmatrix}
\begin{bmatrix}
\phi\\
t_1\\
t_2\\
\vdots\\
t_K
\end{bmatrix}\right)
\end{aligned}
\end{equation}
It is obvious that the Adjoint representation of $SE_K(3)$ can be rewritten as:
\begin{equation}\label{adjoint_representation_1}
Ad_T=\begin{bmatrix}
R&0&0&\cdots&0\\
(p_1)_{\wedge}R&R&0&\cdots&0\\
(p_2)_{\wedge}R&0&R&\cdots&0\\
\vdots&\vdots&\vdots&\ddots&\vdots\\
(p_K)_{\wedge}R&0&0&\cdots&R
\end{bmatrix}\in\mathbb{R}^{3(K+1)\times3(K+1)}
\end{equation}

$Ad_T$ is described as an operator that can act directly on $\mathbb{R}^{3(K+1)}$ instead of on the Lie algebra $\mathfrak{se}_k(3)$.

The set of Adjoint representations of all the elements of $SE_K(3)$ can form a matrix Lie group, we denote it as:
 \begin{equation}\label{adjoint_representation_group}
Ad\left(SE_K(3)\right)=\left\{ \mathcal{T}=Ad_T|T\in SE_K(3) \right\}
 \end{equation}
 
 It turns out that $\left\{ \mathcal{T}=Ad_T|T\in SE_K(3) \right\}
 $ is also a matrix Lie group. For closure we let $\mathcal{T}_1=Ad_{T_1},\mathcal{T}_2=Ad_{T_2}\in Ad(SE_K(3))$, and then 
 \begin{equation}\label{closure_adjoint_group}
 \begin{aligned}
 \mathcal{T}_1\mathcal{T}_2&=Ad_{T_1}Ad_{T_2}=\begin{bmatrix}
 R_1&0&0&\cdots&0\\
 ({p_1}_1)_{\wedge}R_1&R_1&0&\cdots&0\\
 ({p_2}_1)_{\wedge}R_1&0&R_1&\cdots&0\\
 \vdots&\vdots&\vdots&\ddots&\vdots\\
 ({p_K}_1)_{\wedge}R_1&0&0&\cdots&R_1
 \end{bmatrix}\begin{bmatrix}
 R_2&0&0&\cdots&0\\
 ({p_1}_2)_{\wedge}R_2&R_2&0&\cdots&0\\
 ({p_2}_2)_{\wedge}R_2&0&R_2&\cdots&0\\
 \vdots&\vdots&\vdots&\ddots&\vdots\\
 ({p_K}_2)_{\wedge}R_2&0&0&\cdots&R_2
 \end{bmatrix}\\
 &=\begin{bmatrix}
 R_1R_2&0&0&\cdots&0\\
 ({p_1}_1)_{\wedge}R_1R_2+R_1({p_1}_2)_{\wedge}R_2&R_1R_2&0&\cdots&0\\
 ({p_2}_1)_{\wedge}R_1R_2+R_1({p_2}_2)_{\wedge}R_2&0&R_1R_2&\cdots&0\\
 \vdots&\vdots&\vdots&\ddots&\vdots\\
 ({p_K}_1)_{\wedge}R_1R_2+R_1({p_K}_2)_{\wedge}R_2&0&0&\cdots&R_1R_2
 \end{bmatrix}\\
 &=\begin{bmatrix}
 R_1R_2&0&0&\cdots&0\\
 (R_1{p_1}_2+ {p_1}_1)_{\wedge}R_1R_2&R_1R_2&0&\cdots&0\\
 (R_1{p_2}_2+{p_2}_1)_{\wedge}R_1R_2&0&R_1R_2&\cdots&0\\
 \vdots&\vdots&\vdots&\ddots&\vdots\\
 (R_1{p_K}_2+{p_K}_1)_{\wedge}R_1R_2&0&0&\cdots&R_1R_2
 \end{bmatrix}=Ad_{T_1T_2}\in Ad(SE_K(3))
 \end{aligned}
 \end{equation}
 where we have used the linearity of the $\wedge$ operator. Associativity follows from basic properties of matrix multiplication, and the identity element of the group is the $3(K+1)\times 3(K+1)$ identity matrix. 
 For invertibility, we let $\mathcal{T}=Ad_T\in Ad(SE_K(3))$, and then we have
 \begin{equation}\label{adjoint_group_inverse}
 \begin{aligned}
 \mathcal{T}^{-1}&=(Ad_T)^{-1}=\begin{bmatrix}
 R&0&0&\cdots&0\\
 (p_1)_{\wedge}R&R&0&\cdots&0\\
 (p_2)_{\wedge}R&0&R&\cdots&0\\
 \vdots&\vdots&\vdots&\ddots&\vdots\\
 (p_K)_{\wedge}R&0&0&\cdots&R
 \end{bmatrix}^{-1}=
 \begin{bmatrix}
 R^{-1}&0&0&\cdots&0\\
-R^{-1}(p_1)_{\wedge}&R^{-1}&0&\cdots&0\\
-R^{-1} (p_2)_{\wedge}&0&R^{-1}&\cdots&0\\
 \vdots&\vdots&\vdots&\ddots&\vdots\\
-R^{-1} (p_K)_{\wedge}&0&0&\cdots&R^{-1}
 \end{bmatrix}\\
 &= \begin{bmatrix}
 R^{-1}&0&0&\cdots&0\\
 (-R^{-1}p_1)_{\wedge}R^{-1}&R^{-1}&0&\cdots&0\\
  (-R^{-1}p_2)_{\wedge}R^{-1}&0&R^{-1}&\cdots&0\\
 \vdots&\vdots&\vdots&\ddots&\vdots\\
 (-R^{-1}p_K)_{\wedge}R^{-1}&0&0&\cdots&R^{-1}
 \end{bmatrix}=Ad_{T^{-1}}\in Ad(SE_K(3))
 \end{aligned}
 \end{equation}
 
 In the end, it is shown that $Ad(SE_K(3))$ is a matrix Lie group. The Lie group $Ad(SE_K(3))$ plays important role in the invariant filter theory\cite{barrau2016invariant} when the error needs to be switched between the left and right error forms\cite{Ross2020Contact}.
 Meanwhile, we can also consider the adjoint of an element of the Lie algebra $\mathfrak{se}_k(3)$. Let $S=\mathcal{L}(\xi)\in \mathfrak{se}_k(3)$, then the adjoint of this element is $\mathfrak{L}(\xi)$ which has been defined in equation (\ref{adjoint_represnetation_algebra});
 \begin{equation}\label{adjoint_representation_algebra_definition}
\mathfrak{T}:=  ad_S=ad_{\mathcal{L}(\xi)}=\mathfrak{L}(\xi)
 \end{equation}
 
 It is easy to verify that 
 \begin{equation}\label{antisymetry}
 \mathfrak{L}(\xi_1)\xi_2=-\mathfrak{L}(\xi_2)\xi_1, \forall \xi_1,\xi_2\in\mathbb{R}^{3(K+1)}
 \end{equation}
 
 The Lie algebra associated with the matrix Lie group $Ad(SE_K(3))$ is given by
 \begin{equation*}
 \begin{aligned}
 \text{vectorspace}:&\quad ad(\mathfrak{se}_3(k))=\left\{\mathfrak{T}=ad_S\in \mathbb{R}^{3(K+1)\times 3(K+1)}|S \in \mathfrak{se}_k(3)\right\}\\
 \text{field}:&\quad \mathbb{R}\\
 \text{Lie bracket}:&\quad [\mathfrak{T}_1,\mathfrak{T}_2]=\mathfrak{T}_1\mathfrak{T}_2-\mathfrak{T}_2\mathfrak{T}_1
 \end{aligned}
 \end{equation*}
 
 We will omit to show that $ad(\mathfrak{se}_k(3))$ is a vectorspace, but will briefly show that the closure, bilinearity, alternating and Jacaobi identity are satisfied.
 Let $\mathfrak{T},\mathfrak{T}_1=\mathfrak{L}(\xi_1),\mathfrak{T}_2=\mathfrak{L}(\xi_2)\in ad(\mathfrak{se}_k(3))$. Then, for the closure property, we have 
 \begin{equation}\label{closure_property_adjoint_algebra}
 [\mathfrak{T}_1,\mathfrak{T}_2]=\mathfrak{T}_1\mathfrak{T}_2-\mathfrak{T}_2\mathfrak{T}_1=\mathfrak{L}(\xi_1)\mathfrak{L}(\xi_2)-\mathfrak{L}(\xi_2)\mathfrak{L}(\xi_1)
 =\mathfrak{L}\left(\mathfrak{L}(\xi_1) \xi_2\right)
 \in ad\left(\mathfrak{se}_k(3)\right)
 \end{equation}
 
 Bilinearity follows directly from the fact that $\mathfrak{L}(\cdot)$ is a linear operator. The alternating property can be seen easily through
 \begin{equation}\label{alternating_property_adjoint_algebra}
 [\mathfrak{T},\mathfrak{T}]=\mathfrak{T}\mathfrak{T}-\mathfrak{T}\mathfrak{T}=\bf{0}_{3(K+1)}
 \in ad\left(\mathfrak{se}_k(3)\right)
 \end{equation}
 
 Finally, the Jacobi identity can be verified by substituting and applying
 the definition of the Lie bracket. Again, we will refer to $ad(\mathfrak{se}_k(3))$ as the
 Lie algebra, although technically this is only the associated vectorspace.
 
Once again, the Lie algebra $ad(\mathfrak{se}_k(3))$ can be mapped to the Lie group $Ad(SE_K(3))$ through the exponential map as shown below:
\begin{equation}\label{exponential_adjoint_group}
 \begin{aligned}
\mathcal{T}&=\exp_G(\mathfrak{L}(\xi))=\exp_G(\mathfrak{T})=\sum_{n=0}^{\infty}\frac{1}{n!}(\mathfrak{L}(\xi))^n=\sum_{n=0}^{\infty}\frac{1}{n!}\begin{bmatrix}
\phi_{\wedge}&0&0&\cdots&0\\
(t_1)_{\wedge}&\phi_{\wedge}&0&\cdots&0\\
(t_2)_{\wedge}&0&\phi_{\wedge}&\cdots&0\\
\vdots&\vdots&\vdots&\ddots&\vdots\\
(t_K)_{\wedge}&0&0&\cdots&\phi_{\wedge}
\end{bmatrix}^n\\
&=\begin{bmatrix}
R&0&0&\cdots&0\\
(p_1)_{\wedge}R&R&0&\cdots&0\\
(p_2)_{\wedge}R&0&R&\cdots&0\\
\vdots&\vdots&\vdots&\ddots&\vdots\\
(p_K)_{\wedge}R&0&0&\cdots&R
\end{bmatrix}=\begin{bmatrix}
R&0&0&\cdots&0\\
(J_lt_1)_{\wedge}R&R&0&\cdots&0\\
(J_lt_2)_{\wedge}R&0&R&\cdots&0\\
\vdots&\vdots&\vdots&\ddots&\vdots\\
(J_lt_K)_{\wedge}R&0&0&\cdots&R
\end{bmatrix}=\sum_{n=0}^{\infty}\frac{1}{n!}(ad_{\mathcal{L}(\xi)})^n
\end{aligned}
\end{equation}
where $\mathcal{T}\in Ad(SE_K(3)),\mathfrak{L}(\xi)\in ad(\mathfrak{se}_k(3))$ and the key property which has been proven in \cite{barfoot2017state} is used:
\begin{equation}\label{equivalence_aojoint}
(J_lt_k)_{\wedge}R=\sum_{n=0}^{\infty}\sum_{m=0}^{\infty}\frac{1}{(n+m+1)!}(\theta_{\wedge})^n(t_k)_{\wedge}(\theta_{\wedge})^m,k=1,\cdots,K
\end{equation}

For the exponential map of $SE_K(3)$, it can be shown that
\begin{equation}\label{SE_K_3_exponential}
\mathcal{L}(\mathcal{T}\xi)=T\mathcal{L}(\xi)T^{-1}
\end{equation}
\begin{equation}\label{SE_K_3_Ad_exponential}
\mathfrak{L}(\mathcal{T}\xi)=\mathcal{T}\mathfrak{L}(\xi)\mathcal{T}^{-1}
\end{equation}
with $\xi\in\mathbb{R}^{3(K+1)}$, $T\in SE_K(3)$, $\mathcal{T}\in Ad(SE_K(3))$ to establish
\begin{equation}\label{exponential_identities}
T_1T_2(\xi_2)T_1^{-1}=T_2(\mathcal{T}_1\xi_2)
\end{equation}
\begin{equation}\label{exponential_Ad_identities}
\mathcal{T}_1\mathcal{T}_2(\xi_2)\mathcal{T}_1^{-1}=\mathcal{T}_2(\mathcal{T}_1\xi_2)
\end{equation}
where the proofs are similar to equation (\ref{rotation_rotation}).

By equation (\ref{SE_K_3_exponential}), it is also easy to get 
\begin{equation}\label{ad_exp_TxT}
\begin{aligned}
&\exp_G(\mathcal{L}(\mathcal{T}\xi))=\sum_{n=0}^{\infty}\frac{1}{n!}(\mathcal{L}(\mathcal{T}\xi))^n=\sum_{n=0}^{\infty}\frac{1}{n!}(T\mathcal{L}(\xi)T^{-1})^n\\
=&T\sum_{n=0}^{\infty}\frac{1}{n!} (\mathcal{L}(\xi))^nT^{-1}
=T\exp_G(\mathcal{L}(\xi))T^{-1}
\end{aligned}
\end{equation}
This formulation is the same as the equation (\ref{adjoint_representation_definition}).

By equation (\ref{SE_K_3_Ad_exponential}), it is also easy to get 
\begin{equation}\label{Ad_exp_TxT}
\begin{aligned}
&\exp_G(\mathfrak{L}(\mathcal{T}\xi))=\sum_{n=0}^{\infty}\frac{1}{n!}(\mathfrak{L}(\mathcal{T}\xi))^n=\sum_{n=0}^{\infty}\frac{1}{n!}(\mathcal{T}\mathfrak{L}(\xi)\mathcal{T}^{-1})^n\\
=&\mathcal{T}\sum_{n=0}^{\infty}\frac{1}{n!} (\mathfrak{L}(\xi))^n\mathcal{T}^{-1}
=\mathcal{T}\exp_G(\mathfrak{L}(\xi))\mathcal{T}^{-1}
\end{aligned}
\end{equation}

The logarithm mapping is defined through:
\begin{equation}\label{logarithm_mapping_adjoint}
\xi=\mathfrak{L}^{-1}(\log_G(\mathcal{T}))
\end{equation}

The exponential mapping is again surjective-only: every $\mathcal{T}\in Ad(SE_K(3))$ can be generated by many $\xi\in \mathbb{R}^{3(K+1)}$.

So far, we see a nice commutative relationship between the various Lie groups and algebras associated with $SE_K(3)$:
\begin{displaymath}
\xymatrix{
	&\text{Lie algebra} & \text{Lie group}\\
(K+3)\times (K+3) & S=\mathcal{L}(\xi)\in \mathfrak{se}_k(3) \ar[r]^{\exp_G} \ar[d]_{ad}     & T\in SE_K(3) \ar[d]^{Ad} \\
3(K+1)\times 3(K+1)& \mathfrak{T}=\mathfrak{L}(\xi)\in ad(\mathfrak{se}_k(3)) \ar[r]^{\exp_G}       & \mathcal{T}\in Ad(SE_K(3))
}
\end{displaymath}
The two paths from the Lie algebra $\mathfrak{se}_k(3)$ to the Lie group $Ad(SE_K(3))$ amount to show that the Adjoint representation $Ad(SE_K(3))$ is the automorphism of the group $SE_K(3)$ and therefore we get
\begin{equation}\label{adjoint_Adjoint}
\underbrace{Ad\left(\exp_G(\mathcal{L}(\xi))\right)}_{\mathcal{T}}=\exp_G\left(\underbrace{ad(\mathcal{L}(\xi))}_{\mathfrak{L}(\xi)}\right)
\end{equation}
which implies that we can go from $\xi\in\mathbb{R}^{3(K+1)}$ to $\mathcal{T}\in Ad(SE_K(3))$ and back.

Similarly to the direct series expression for $T$, we can also work one out for $\mathcal{T}=Ad_T$ by using the identity
\begin{equation}\label{direct_series_adjoint_representation}
\left(\mathfrak{L}(\xi)\right)^5+2\theta^2\left(\mathfrak{L}(\xi) \right)^3+\theta^4 \mathfrak{L}(\xi)=\mathfrak{T}^5+2\theta^2\mathfrak{T}^3+\theta^4\mathfrak{T}=\bf{0}_{3(K+1)}
\end{equation}

Expanding the series and using the identity to rewrite all quintic and higher terms in lower-order terms, we have
\begin{equation}\label{Exponential_map_adjoint_se_k_3}
\begin{aligned}
&\quad \mathcal{T} =Exp(\xi)=\exp_G(\mathfrak{L}(\xi))=\exp_G(\mathfrak{T})=\sum_{n=0}^{\infty}\frac{1}{n!}\mathfrak{T}^n\\
&=I_{3(K+1)}+\mathfrak{T}+\frac{1}{2!}\mathfrak{T}^2+\frac{1}{3!}\mathfrak{T}^3+\frac{1}{4!}\mathfrak{T}^4+\frac{1}{5!}\mathfrak{T}^5+\cdots\\
&=I_{3(K+1)}+\underbrace{\left(1-\frac{1}{5!}\theta^4+\frac{2}{7!}\theta^6-\frac{3}{9!}\theta^8+\frac{4}{11!}\theta^{10}-\cdots \right)}_{\frac{3\sin\theta-\theta\cos\theta}{2\theta}}\mathfrak{T}\\
&\quad +\underbrace{\left(\frac{1}{2!}-\frac{1}{6!}\theta^4+\frac{2}{8!}\theta^6-\frac{3}{10!}\theta^8+\frac{4}{12!}\theta^{10}-\cdots \right)}_{\frac{4-\theta\sin\theta-4\cos\theta}{2\theta^2}} \mathfrak{T}^2\\
&\quad +\underbrace{\left(\frac{1}{3!}-\frac{2}{5!}\theta^2+\frac{3}{7!}\theta^4-\frac{4}{9!}\theta^6+\frac{4}{11!}\theta^{8}-\cdots \right)}_{\frac{\sin\theta-\theta\cos\theta}{2\theta^3}} \mathfrak{T}^3\\
&\quad +\underbrace{\left(\frac{1}{4!}-\frac{2}{6!}\theta^2+\frac{3}{8!}\theta^4-\frac{4}{10!}\theta^6+\frac{5}{12!}\theta^{8}-\cdots \right)}_{\frac{2-\theta\sin\theta-2\cos\theta}{2\theta^4}} \mathfrak{T}^4\\
&=I_{3(K+1)}+\left(\frac{3\sin\theta-\theta\cos\theta}{2\theta}\right)\mathfrak{T}+\left(\frac{4-\theta\sin\theta-4\cos\theta}{2\theta^2}\right)\mathfrak{T}^2\\
&\quad +\left(\frac{\sin\theta-\theta\cos\theta}{2\theta^3}\right)\mathfrak{T}^3+\left(\frac{2-\theta\sin\theta-2\cos\theta}{2\theta^4}\right)\mathfrak{T}^4\in\mathbb{R}^{3(K+1)\times3(K+1)}
\end{aligned}
\end{equation}
\subsection{Actions of the group $SE_K(3)$}
Each group $SE_K(3)$ comes with a family of natural group actions allowing to recover many of the observation functions of robotics and navigation. 
The vector action of the group $SE_K(3)$ on $\mathbb{R}^3$ with parameters $(\gamma_1,\cdots,\gamma_K)^T\in \mathbb{R}^K$ can be defined as:
\begin{equation}\label{group_action}
x \circ b= Rb+\sum_{i=1}^{K}\gamma_i r_i \in\mathbb{R}^3,x=\exp_G(\mathcal{L}(R,r_1,\cdots,r_K))\in SE_K(3),b\in\mathbb{R}^3
\end{equation}

It is easy to check that this defines an action of the group $SE_K(3)$:
Given $x=\exp_G(\mathcal{L}(R,r_1,\cdots,r_K))\in SE_K(3)$, $x'=\exp_G(\mathcal{L}(R',t_1,\cdots,t_K))\in SE_K(3)$, and $b\in\mathbb{R}^3$, we have:
\begin{equation}\label{verify_group_action}
\begin{aligned}
x\circ (x'\circ b)&=x\circ (R'b+\sum_{i=1}^{K}\gamma_i t_i)=R(R'b+\sum_{i=1}^{K}\gamma_i t_i)+\sum_{i=1}^{K}\gamma_i r_i\\
&=RR'b+\sum_{i=1}^{K}\gamma_i(Rt_i+r_i)=(x x')\circ b
\end{aligned}
\end{equation}
where we used the group multiplication $xx'=(RR',Rt_1+r_1,\cdots,Rt_K+r_K)$.
\subsection{Baker-Campbell-Hausdorff}
Given two non-commuting elements $A, B\in \mathfrak{se}_k(3)$, the well known Baker-Campbell-Hausdorff (BCH) formula states that the element $C\in \mathfrak{se}_k(3)$ defined by $\exp_G(A)\exp_G(B)=\exp_G(C)$ can be expressed in the vectorspace without requiring the application of the exponential or the logarithm map:
\begin{equation}\label{BCH_group}
\begin{aligned}
C&=\log_G(\exp_G(A)\exp_G(B))\\
&=A+B+\frac{1}{2}[A,B]+\frac{1}{12}([A,[A,B]]+[B,[B,A]])-\frac{1}{24}[B,[A,[A,B]]]+\cdots
\end{aligned}
\end{equation}

In particular cases of $SE_K(3)$ and $Ad(SE_K(3))$, let $\exp_G(\mathcal{L}(\xi_1)),\exp_G(\mathcal{L}(\xi_2))\in SE_K(3)$, and $ \exp_G(\mathfrak{L}(\xi_1)),\exp_G(\mathfrak{L}(\xi_2))\in Ad(SE_K(3))$, we can show that
\begin{equation}\label{BCH_group_SE_K_3}
\begin{aligned}
&\quad \xi=\mathcal{L}^{-1}(S)=\mathcal{L}^{-1}\left(\log_G(\exp_G(S_1)\exp_G(S_2) \right)=\mathcal{L}^{-1}\left(\log_G(\exp_G(\mathcal{L}(\xi_1))\exp_G(\mathcal{L}(\xi_2))) \right)\\
&=\xi_1+\xi_2+\frac{1}{2!}\mathfrak{L}(\xi_1)\xi_2+\frac{1}{12!}\mathfrak{L}(\xi_1)\mathfrak{L}(\xi_1)\xi_2+\frac{1}{12!}\mathfrak{L}(\xi_2)\mathfrak{L}(\xi_2)\xi_1-\frac{1}{24!}\mathfrak{L}(\xi_2)\mathfrak{L}(\xi_1)\mathfrak{L}(\xi_1)\xi_2+\cdots
\end{aligned}
\end{equation}
\begin{equation}\label{BCH_group_Ad_SE_K_3}
\begin{aligned}
&\quad \xi=\mathfrak{L}^{-1}(\mathfrak{T})=\mathfrak{L}^{-1}\left(\log_G(\exp_G(\mathfrak{T}_1)\exp_G(\mathfrak{T}_2)) \right)=\mathfrak{L}^{-1}\left(\log_G(\exp_G(\mathfrak{L}(\xi_1))\exp_G(\mathfrak{L}(\xi_2))) \right)\\
&=\xi_1+\xi_2+\frac{1}{2!}\mathfrak{L}(\xi_1)\xi_2+\frac{1}{12!}\mathfrak{L}(\xi_1)\mathfrak{L}(\xi_1)\xi_2+\frac{1}{12!}\mathfrak{L}(\xi_2)\mathfrak{L}(\xi_2)\xi_1-\frac{1}{24!}\mathfrak{L}(\xi_2)\mathfrak{L}(\xi_1)\mathfrak{L}(\xi_1)\xi_2+\cdots
\end{aligned}
\end{equation}

Alternatively, if we assume that $\xi_1$ or $\xi_2$ is small, then using the approximate BCH formulas, we can show that
\begin{equation}\label{BCH_group_SE_K_3_Jacobian}
\begin{aligned}
&\quad \xi=\mathcal{L}^{-1}\left(\log_G(\exp_G(\mathcal{L}(\xi_1))\exp_G(\mathcal{L}(\xi_2))) \right)\approx \left\{ \begin{array}{lr}
	\mathcal{J}_l(\xi_2)^{-1}\xi_1+\xi_2& \text{if $\xi_1$ small}\\
	\xi_1+\mathcal{J}_r(\xi_1)^{-1}\xi_2& \text{if $\xi_2$ small}
	\end{array}
	\right.
\end{aligned}
\end{equation}
\begin{equation}\label{BCH_group_Ad_SE_K_3_Jacobian}
\begin{aligned}
&\quad \xi=\mathfrak{L}^{-1}\left(\log_G(\exp_G(\mathfrak{L}(\xi_1))\exp_G(\mathfrak{L}(\xi_2))) \right)\approx \left\{ \begin{array}{lr}
\mathcal{J}_l(\xi_2)^{-1}\xi_1+\xi_2& \text{if $\xi_1$ small}\\
\xi_1+\mathcal{J}_r(\xi_1)^{-1}\xi_2& \text{if $\xi_2$ small}
\end{array}
\right.
\end{aligned}
\end{equation}
where $\mathcal{J}_l(\xi)^{-1}$ and $\mathcal{J}_r(\xi)^{-1}$ are denoted as the inverse of the left and right Jacobians of matrix Lie group $SE_K(3)$, respectively. And they are given as 
\begin{equation}\label{inverse_left_jacobian_gourp}
\mathcal{J}_l(\xi)^{-1}=\sum_{n=0}^{\infty}\frac{B_n}{n!}(\mathfrak{L}(\xi))^n= \sum_{n=0}^{\infty}\frac{B_n}{n!}(ad_{\mathcal{L}(\xi)})^n   =\frac{ad_{\mathcal{L}(\xi)}}{\exp_G(ad_{\mathcal{L}(\xi)})-I}
\end{equation}
\begin{equation}\label{inverse_right_jacobian_gourp}
\mathcal{J}_r(\xi)^{-1}=\sum_{n=0}^{\infty}\frac{B_n}{n!}(-\mathfrak{L}(\xi))^n= \sum_{n=0}^{\infty}\frac{B_n}{n!}(-ad_{\mathcal{L}(\xi)})^n   =\frac{ad_{\mathcal{L}(\xi)}}{-\exp_G(-ad_{\mathcal{L}(\xi)})+I}
\end{equation}

On the other hand, if we perturb the Lie algebra by adding small element $\xi$, which is equivalent to multiplying the Lie group $SE_K(3)$ and $Ad(SE_K(3))$ by small increments, we can get the left and right form of BCH formula separately
\begin{equation}\label{lie_algebra_perturb_left}
\exp_G(\mathcal{L}(\delta\xi+\xi))=\exp_G(\mathcal{L}(\mathcal{J}_l(\xi)^{-1}\mathcal{J}_l(\xi)\delta\xi+\xi))\approx \exp_G(\mathcal{L}(\mathcal{J}_l(\xi)\delta\xi))\exp_G(\mathcal{L}(\xi))
\end{equation}
\begin{equation}\label{lie_algebra_perturb_right}
\exp_G(\mathcal{L}(\xi+\delta\xi))=\exp_G(\mathcal{L}(\xi+\mathcal{J}_r(\xi)^{-1}\mathcal{J}_r(\xi)\delta\xi))\approx \exp_G(\mathcal{L}(\xi))\exp_G(\mathcal{L}(\mathcal{J}_r(\xi)\delta\xi))
\end{equation}
\begin{equation}\label{lie_algebra_perturb_left_Ad}
\exp_G(\mathfrak{L}(\delta\xi+\xi))=\exp_G(\mathfrak{L}(\mathcal{J}_l(\xi)^{-1}\mathcal{J}_l(\xi)\delta\xi+\xi))\approx \exp_G(\mathfrak{L}(\mathcal{J}_l(\xi)\delta\xi))\exp_G(\mathfrak{L}(\xi))
\end{equation}
\begin{equation}\label{lie_algebra_perturb_right_Ad}
\exp_G(\mathfrak{L}(\xi+\delta\xi))=\exp_G(\mathfrak{L}(\xi+\mathcal{J}_r(\xi)^{-1}\mathcal{J}_r(\xi)\delta\xi))\approx \exp_G(\mathfrak{L}(\xi))\exp_G(\mathfrak{L}(\mathcal{J}_r(\xi)\delta\xi))
\end{equation}

The expressions for the right and left Jacobians of $SE_K(3)$ can be given as 
\begin{equation}\label{left_jacobian_gourp}
\begin{aligned}
\mathcal{J}_l(\xi)&=\sum_{n=0}^{\infty}\frac{1}{(n+1)!}(\mathfrak{L}(\xi))^n=\int_{0}^{1}\mathcal{T}^{\alpha}d\alpha=\begin{bmatrix}
J_l   &0     &0   &\cdots & 0\\
{Q_1}_l &J_l   &0   &\cdots &0\\
{Q_2}_l &0     &J_l &\cdots &0\\
\vdots&\vdots&\vdots&\ddots&\vdots\\
{Q_K}_l &0     &0   &\cdots &J_l
\end{bmatrix}\\
&=\sum_{n=0}^{\infty}\frac{1}{(n+1)!}(ad_{\mathcal{L}(\xi)})^n=\frac{\exp_G(ad_{\mathcal{L}(\xi)})-I}{ad_{\mathcal{L}(\xi)}}
\end{aligned}
\end{equation}
\begin{equation}\label{right_jacobian_gourp}
\begin{aligned}
\mathcal{J}_r(\xi)&=\sum_{n=0}^{\infty}\frac{1}{(n+1)!}(-\mathfrak{L}(\xi))^n=\int_{0}^{1}\mathcal{T}^{-\alpha}d\alpha=\begin{bmatrix}
J_r   &0     &0   &\cdots & 0\\
{Q_1}_r &J_r   &0   &\cdots &0\\
{Q_2}_r &0     &J_r &\cdots &0\\
\vdots&\vdots&\vdots&\ddots&\vdots\\
{Q_K}_r &0     &0   &\cdots &J_r
\end{bmatrix}\\
&=\sum_{n=0}^{\infty}\frac{1}{(n+1)!}(-ad_{\mathcal{L}(\xi)})^n=\frac{\exp_G(-ad_{\mathcal{L}(\xi)})-I}{-ad_{\mathcal{L}(\xi)}}
=\frac{-\exp_G(-ad_{\mathcal{L}(\xi)})+I}{ad_{\mathcal{L}(\xi)}}
\end{aligned}
\end{equation}
where
\begin{equation}\label{Q_k_l}
\begin{aligned}
{Q_k}_l(\xi)&=\sum_{n=0}^{\infty}\sum_{m=0}^{\infty}(\phi_{\wedge})^n(t_k)_{\wedge}(\phi_{\wedge})^m\\
&=\frac{1}{2}(t_k)_{\wedge}+(\frac{\theta-\sin\theta}{\theta^3})(\phi_{\wedge}(t_k)_{\wedge}+(t_k)_{\wedge}\phi_{\wedge}+\phi_{\wedge}(t_k)_{\wedge}\phi_{\wedge})\\
&\qquad +(\frac{\theta^2+2\cos\theta-2}{2\theta^4})(\phi_{\wedge}\phi_{\wedge}(t_k)_{\wedge}+(t_k)_{\wedge}\phi_{\wedge}\phi_{\wedge}-3\phi_{\wedge}(t_k)_{\wedge}\phi_{\wedge})\\
&\qquad +(\frac{2\theta-3\sin\theta+\theta\cos\theta}{2\theta^5})(\phi_{\wedge}(t_k)_{\wedge}\phi_{\wedge}\phi_{\wedge}+\phi_{\wedge}\phi_{\wedge}(t_k)_{\wedge}\phi_{\wedge})
\end{aligned}
\end{equation}
\begin{equation}\label{right_Jacobian}
\begin{aligned}
&J_r=J_r(\phi)=\sum_{n=0}^{\infty}\frac{1}{(n+1)!}(-\phi_{\wedge})^n=\int_{0}^{1}R^{-\alpha}d\alpha=I_3-\frac{1-\cos\theta}{\theta}u_{\wedge}+(1-\frac{\sin\theta}{\theta})u_{\wedge}^2\\
=&I_3-\frac{1-\cos\theta}{\theta}u_{\wedge}+\frac{\theta-\sin\theta}{\theta}(-I_3+uu^T)
=\frac{\sin\theta}{\theta}I_3-\frac{1-\cos\theta}{\theta}u_{\wedge}+(1-\frac{\sin\theta}{\theta})uu^T\\
=&I_3-\frac{1-\cos\theta}{\theta^2}\phi_{\wedge}+\frac{\theta-\sin\theta}{\theta^3}\phi_{\wedge}^2
\end{aligned}
\end{equation}
$\mathcal{T}=\exp_G(\mathfrak{L}(\xi))$, $R=\exp_G(\phi_{\wedge})$,  and $\xi=\begin{bmatrix}
\phi^T &t_1^T&t_2^T& \cdots & t_K^T
\end{bmatrix}^T \in \mathbb{R}^{3(K+1)}$. $J_l$ and $J_r$ are referred to as the right and left Jacobians of $SO(3)$, respectively. The expression for ${Q_k}_l$ comes from expanding the series and grouping terms into
the series forms of the trigonometric functions\cite{barfoot2017state}.

The inverse for the right Jacobian matrix $J_r$ is
\begin{equation}\label{inverse_right_Jacobian_matrix}
	\begin{aligned}
	&J_r^{-1}=	J_r^{-1}(\phi)=\sum_{n=0}^{\infty}\frac{B_n}{n!}(-\phi_{\wedge})^n=I_3+\frac{\theta}{2}u_{\wedge}+(1-\frac{\theta}{2}\cot\frac{\theta}{2})u_\wedge u_{\wedge}\\
		=&I_3+\frac{\theta}{2}u_{\wedge}+(1-\frac{\theta}{2}\cot\frac{\theta}{2})(-I_3+uu^T)
		=\frac{\theta}{2}\cot\frac{\theta}{2}I_3+(1-\frac{\theta}{2}\cot\frac{\theta}{2})uu^T+\frac{\theta}{2}u_{\wedge}\\
		=&I+\frac{1}{2}\phi_{\wedge}+\left(\frac{1}{\theta^2}-\frac{1+\cos\theta}{2\theta\sin\theta} \right)\phi_{\wedge}^2
	\end{aligned}
\end{equation}

The identity relationship between the Rodriguez formula and the right Jacobian matrix can be proofed as following:
\begin{equation}\label{right_jacobian_rodriguez}
\begin{aligned}
J_r(R\phi)&=I_3-\frac{1-\cos\theta}{\theta^2}(R\phi)_{\wedge}+\frac{\theta-\sin\theta}{\theta^3}(R\phi)_{\wedge}^2\\
&=I_3-\frac{1-\cos\theta}{\theta^2}R\phi_{\wedge}R^T+\frac{\theta-\sin\theta}{\theta^3}R\phi_{\wedge}R^TR\phi_{\wedge}R^T\\
&=RR^T-\frac{1-\cos\theta}{\theta^2}R\phi_{\wedge}R^T+\frac{\theta-\sin\theta}{\theta^3}R\phi_{\wedge}^2R^T=RJ_r(\phi)R^T
\end{aligned}
\end{equation}
\begin{equation}\label{right_jacobian_rodriguez_2}
\begin{aligned}
&J_r(\phi)\phi_{\wedge}=\left(I_3-\frac{1-\cos\theta}{\theta^2}\phi_{\wedge}+\frac{\theta-\sin\theta}{\theta^3}\phi_{\wedge}^2\right)\phi_{\wedge}=\phi_{\wedge}-\frac{1-\cos\theta}{\theta^2}\phi_{\wedge}^2+\frac{\theta-\sin\theta}{\theta^3}\phi_{\wedge}^3\\
=&\phi_{\wedge}-\frac{1-\cos\theta}{\theta^2}\phi_{\wedge}^2+(\theta-\sin\theta)u_{\wedge}^3=\phi_{\wedge}-\frac{1-\cos\theta}{\theta^2}\phi_{\wedge}^2+(\theta-\sin\theta)(-u_{\wedge})\\
=&-\frac{1-\cos\theta}{\theta^2}\phi_{\wedge}^2+\frac{\sin\theta}{\theta}\phi_{\wedge}=I_3-R(-\phi)
\end{aligned}
\end{equation}

The relations for ${Q_k}_r$ come from the relationships between the left and right Jacobians:
\begin{equation}\label{Jacobian_left_from_right}
\mathcal{J}_l(\xi)=\mathcal{T}\mathcal{J}_r(\xi)=Ad_{T}\mathcal{J}_r(\xi)=Ad_{Exp(\xi)}\mathcal{J}_r(\xi)
\end{equation}
\begin{equation}\label{Jacobian_left_to_right}
\mathcal{J}_l(-\xi)=\mathcal{J}_r(\xi)
\end{equation}

Equation (\ref{Jacobian_left_from_right}) is proved by 
\begin{equation}\label{Jacobian_left_from_right_prove}
\mathcal{T}\mathcal{J}_r(\xi)=\mathcal{T}\int_{0}^{1}\mathcal{T}^{-\alpha}d\alpha=\int_{0}^{1}\mathcal{T}^{1-\alpha}d\alpha=-\int_{1}^{0}\mathcal{T}^{\beta}d\beta=\int_{0}^{1}\mathcal{T}^{\beta}d\beta=\mathcal{J}_l(\xi)
\end{equation}

Equation (\ref{Jacobian_left_to_right}) is to be true from
\begin{equation}\label{Jacobian_left_to_right_prove}
\mathcal{J}_r(\xi)=\int_{0}^{1}\mathcal{T}(\xi)^{-\alpha}d\alpha=\int_{0}^{1}(\mathcal{T}(\xi)^{-1})^{\alpha}d\alpha=\int_{0}^{1}(\mathcal{T}(-\xi))^{\alpha}d\alpha=\mathcal{J}_l(-\xi)
\end{equation}

From the equation (\ref{Jacobian_left_from_right}) we have
\begin{equation}\label{Jacobian_left_from_right_deduce}
J_l=RJ_r,{Q_k}_l=R{Q_k}_r+(J_lt_k)_{\wedge}RJ_r,k=1,\cdots,K
\end{equation}

From the equation (\ref{Jacobian_left_to_right}) we have
\begin{equation}\label{Jacobian_left_to_right_deduce}
J_l(-\xi)=J_r(\xi),{Q_k}_l(-\xi)={Q_k}_r(\xi),k=1,\cdots,K
\end{equation}

From the Taylor series expressions of $\mathcal{T}$ and $\mathcal{J}_l$, we know that
\begin{equation}\label{T_and_Jacobian_left}
\mathcal{T}=\sum_{n=0}^{\infty}\frac{1}{n!}(\mathfrak{L}(\xi))^n=I_{3(K+1)}+\mathfrak{L}(\xi)\sum_{n=0}^{\infty}\frac{1}{(n+1)!}(\mathfrak{L}(\xi))^n=I_{3(K+1)}+\mathfrak{L}(\xi)\mathcal{J}_l
\end{equation}

Expanding the expression for the Jacobian, we have that
\begin{equation}\label{Jacobian_Ad_group}
\mathcal{J}_l(\xi)=\sum_{n=0}^{\infty}\frac{1}{(n+1)!}(\mathfrak{L}(\xi))^n=I_{3(K+1)}+\alpha_1\mathfrak{L}(\xi)+\alpha_2(\mathfrak{L}(\xi))^2+\alpha_3(\mathfrak{L}(\xi))^3+\alpha_4(\mathfrak{L}(\xi))^4
\end{equation}
where $\alpha_1, \alpha_2, \alpha_3, \alpha_4$ are unknown coefficients. These series can be expressed using only terms up to quartic through the use of the identity in equation (\ref{direct_series_adjoint_representation}). Substituting this into equation (\ref{T_and_Jacobian_left}) we see that
\begin{equation}\label{T_and_Jacobian_left_series}
\mathcal{T}=I_{3(K+1)}+\mathfrak{L}(\xi)+\alpha_1(\mathfrak{L}(\xi))^2+\alpha_2(\mathfrak{L}(\xi))^3+\alpha_3(\mathfrak{L}(\xi))^4+\alpha_4(\mathfrak{L}(\xi))^5
\end{equation}

Using equation (\ref{direct_series_adjoint_representation}) to rewrite the quintic term using the lower-order terms, we have that
\begin{equation}\label{T_and_Jacobian_left_series_lower_order_terms}
\mathcal{T}=I_{3(K+1)}+(\frac{1}{1!}-\alpha_4\theta^4)\mathfrak{L}(\xi)+\alpha_1(\mathfrak{L}(\xi))^2+(\alpha_2-2\alpha_4\theta^2)(\mathfrak{L}(\xi))^3+\alpha_3(\mathfrak{L}(\xi))^4
\end{equation}

Comparing the coefficients to those in equation (\ref{Exponential_map_adjoint_se_k_3}), we can solve for $\alpha_1$, $\alpha_2$, $\alpha_3$ and $\alpha_4$ such that
\begin{equation}\label{T_and_Jacobian_left_series_lower_order_terms_analytic}
\begin{aligned}
 \mathcal{J}_l(\xi)&=I_{3(K+1)}+\left(\frac{4-\theta\sin\theta-4\cos\theta}{2\theta^2}\right)\mathfrak{T}+\left(\frac{4\theta-5\sin\theta+\theta\cos\theta}{2\theta^3}\right)\mathfrak{T}^2\\
&\quad +\left(\frac{2-\theta\sin\theta-2\cos\theta}{2\theta^4}\right)\mathfrak{T}^3+\left(\frac{2\theta-3\sin\theta+\theta\cos\theta}{2\theta^5}\right)\mathfrak{T}^4\in\mathbb{R}^{3(K+1)\times3(K+1)}
\end{aligned}
\end{equation}

This avoids the need to work out $J_l$ and ${Q_k}_l, k=1,\cdots,K$  individually and then to assemble them into $\mathcal{J}_l$.

For some applications involving linearization and discretization, the following auxiliary quantity is introduced:
  \begin{equation}\label{auxilary_function_general}
  \Gamma_{m}(\xi)=\sum_{n=0}^{\infty}\frac{1}{(n+m)!}(\mathfrak{L}(\xi))^n=\sum_{n=0}^{\infty}\frac{1}{(n+m)!}\mathfrak{T}^n
  \end{equation}
It is obvious that it draws on the series expansion of the matrix exponential. For $m=0$, it yields:
\begin{equation}\label{gamma_0}
  \Gamma_{0}(\xi)=\sum_{n=0}^{\infty}\frac{1}{(n)!}(\mathfrak{L}(\xi))^n=\mathcal{T}
\end{equation}

For $m=1$, it yields
\begin{equation}\label{gamma_1}
\Gamma_{1}(\xi)=\sum_{n=0}^{\infty}\frac{1}{(n+1)!}(\mathfrak{L}(\xi))^n=\mathcal{J}_l(\xi)
\end{equation}

For $m=2$, it yields
\begin{equation}\label{gamma_2}
\Gamma_{2}(\xi)=\sum_{n=0}^{\infty}\frac{1}{(n+2)!}(\mathfrak{L}(\xi))^n
\end{equation}

In order to obtain the closed-form expression for $\Gamma_{2}(\xi)$, the similar procedure as in derivation $\Gamma_{1}(\xi)$ id adopted. The relationship between $\Gamma_{1}(\xi)$ and $\Gamma_{2}(\xi)$ can be expressed as
\begin{equation}\label{gamma_2_derivation}
\begin{aligned}
&\Gamma_{1}(\xi)= I_{3(K+1)}+\Gamma_{2}(\xi)\mathfrak{L}(\xi)\\
=&I_{3(K+1)}+(\frac{1}{2!}-\alpha_4\theta^4)\mathfrak{L}(\xi)+\alpha_1(\mathfrak{L}(\xi))^2+(\alpha_2-2\alpha_4\theta^2)(\mathfrak{L}(\xi))^3+\alpha_3(\mathfrak{L}(\xi))^4
\end{aligned}
\end{equation}

Comparing equation (\ref{gamma_2_derivation}) with equation (\ref{T_and_Jacobian_left_series_lower_order_terms_analytic}), the coefficients of $\Gamma_{2}(\xi)$ can be obtained, that is
\begin{equation}\label{coefficient_gamma_2}
\begin{aligned}
\alpha_1=\left(\frac{4\theta-5\sin\theta+\theta\cos\theta}{2\theta^3}\right),
\alpha_2=\left( \frac{2\theta^2-6+\theta\sin\theta+6\cos\theta}{2\theta^4}\right),\\
\alpha_3=\left(\frac{2\theta-3\sin\theta+\theta\cos\theta}{2\theta^5}\right),
\alpha_4=\left(  \frac{\theta^2-4+\theta\sin\theta+4\cos\theta}{2\theta^6} \right)
\end{aligned}
\end{equation}
Therefore, the closed-form expression for $\Gamma_{2}(\xi)$ is 
\begin{equation}\label{gamma_2_final}
\begin{aligned}
\Gamma_{2}(\xi)=\frac{1}{2!}\cdot I_{3(K+1)}+\left(\frac{4\theta-5\sin\theta+\theta\cos\theta}{2\theta^3}\right)\mathfrak{T}+\left( \frac{2\theta^2-6+\theta\sin\theta+6\cos\theta}{2\theta^4}\right)\mathfrak{T}^2\\
+\left(\frac{2\theta-3\sin\theta+\theta\cos\theta}{2\theta^5}\right)\mathfrak{T}^3
+\left(  \frac{\theta^2-4+\theta\sin\theta+4\cos\theta}{2\theta^6} \right)\mathfrak{T}^4
\end{aligned}
\end{equation}

 In order to obtain the closed-form expression for $\Gamma_{3}(\xi)$, the relationship between $\Gamma_{2}(\xi)$ and $\Gamma_{3}(\xi)$ can be given as
 \begin{equation}\label{gamma_3_derivation}
 \begin{aligned}
  &\Gamma_{2}(\xi)=\frac{1}{2!}\cdot I_{3(K+1)}+\Gamma_{3}(\xi)\mathfrak{L}(\xi)\\
 =&\frac{1}{2!}\cdot I_{3(K+1)}+(\frac{1}{3!}-\alpha_4\theta^4)\mathfrak{L}(\xi)+\alpha_1(\mathfrak{L}(\xi))^2+(\alpha_2-2\alpha_4\theta^2)(\mathfrak{L}(\xi))^3+\alpha_3(\mathfrak{L}(\xi))^4
 \end{aligned}
 \end{equation}
 
 Comparing equation (\ref{gamma_3_derivation}) with equation (\ref{gamma_2_final}), the coefficients of $\Gamma_{2}(\xi)$ can be obtained, that is
 \begin{equation}\label{coefficient_gamma_3}
 \begin{aligned}
 \alpha_1&=\left( \frac{2\theta^2-6+\theta\sin\theta+6\cos\theta}{2\theta^4}\right),
 \alpha_2=\left( \frac{2\theta^3-18\theta+21\sin\theta-3\theta\cos\theta }{6\theta^5}\right),\\
 \alpha_3&=\left(  \frac{\theta^2-4+\theta\sin\theta+4\cos\theta}{2\theta^6} \right),
  \alpha_4=\left(\frac{\theta^3-12\theta+15\sin\theta-3\theta\cos\theta}{6\theta^7}\right)
 \end{aligned}
 \end{equation}
 Therefore, the closed-form expression for $\Gamma_{3}(\xi)$ is 
 \begin{equation}\label{gamma_3_final}
 \begin{aligned}
 \Gamma_{3}(\xi)=&\frac{1}{3!}\cdot I_{3(K+1)}+\left( \frac{2\theta^2-6+\theta\sin\theta+6\cos\theta}{2\theta^4}\right)\mathfrak{T}\\
& +\left( \frac{2\theta^3-18\theta+21\sin\theta-3\theta\cos\theta }{6\theta^5}\right)\mathfrak{T}^2\\
& +\left(  \frac{\theta^2-4+\theta\sin\theta+4\cos\theta}{2\theta^6} \right)\mathfrak{T}^3
  +\left(\frac{\theta^3-12\theta+15\sin\theta-3\theta\cos\theta}{6\theta^7}\right)\mathfrak{T}^4
 \end{aligned}
 \end{equation}
 
 Owing to equation (\ref{SE_K_3_Ad_exponential}), we have that
 \begin{equation}\label{adjoint_gamma_m}
 \Gamma_{m}\left(\Gamma_{0}(\xi)\xi\right)=\Gamma_{0}(\xi)\Gamma_{m}(\xi)\Gamma_{0}(\xi)^{-1} \Rightarrow \Gamma_{0}(\xi)\Gamma_{m}(\xi)=\Gamma_{m}\left(\Gamma_{0}(\xi)\xi\right)\Gamma_{0}(\xi)
 \end{equation}
 The proof is as follows:
 \begin{equation}\label{proof_adjoint_gamma_m}
 \begin{aligned}
 &\Gamma_{0}(\xi)\Gamma_{m}(\xi)\Gamma_{0}(\xi)^{-1} \\
 =&\Gamma_{0}(\xi)\left(\alpha_0 I_{3(K+1)}+\alpha_1\mathfrak{L}(\xi)+\alpha_2(\mathfrak{L}(\xi))^2+\alpha_3(\mathfrak{L}(\xi))^3+\alpha_4(\mathfrak{L}(\xi))^4\right)\Gamma_{0}(\xi)^{-1} \\
 =&\alpha_0 I_{3(K+1)}+\alpha_1\Gamma_{0}(\xi)\mathfrak{L}(\xi)\Gamma_{0}(\xi)^{-1}+\alpha_2\Gamma_{0}(\xi)(\mathfrak{L}(\xi))^2\Gamma_{0}(\xi)^{-1}\\
 &+\alpha_3\Gamma_{0}(\xi)(\mathfrak{L}(\xi))^3\Gamma_{0}(\xi)^{-1}+\alpha_4\Gamma_{0}(\xi)(\mathfrak{L}(\xi))^4\Gamma_{0}(\xi)^{-1}\\
 =&\alpha_0 I_{3(K+1)}+\alpha_1\mathfrak{L}(\Gamma_{0}(\xi)\xi)+\alpha_2(\mathfrak{L}(\Gamma_{0}(\xi)\xi))^2
 +\alpha_3(\mathfrak{L}(\Gamma_{0}(\xi)\xi))^3+\alpha_4(\mathfrak{L}(\Gamma_{0}(\xi)\xi))^4\\
 =& \Gamma_{m}\left(\Gamma_{0}(\xi)\xi\right)
 \end{aligned}
 \end{equation}

The linearization of the function $\Gamma_{m}(\cdot)$ is the first order Taylor series of the function evaluated as a certain elements of the domain. If we assume that $\phi$ is small, then using the first order Taylor series we can obtain
\begin{equation}\label{left_jacobian_333}
\Gamma_{m}(\phi+\psi)\approx Exp(\Gamma_{m+1}(\psi)\phi)\Gamma_{m}(\psi)=\Gamma_{0}(\Gamma_{m+1}(\psi)\phi)\Gamma_{m}(\psi)
\end{equation}
The left Jacobian matrix approximation in the BCH formula given in equation (\ref{lie_algebra_perturb_left_Ad}) can also be obtained by equation (\ref{left_jacobian_333}) when $m=0$.

If we assume that $\phi$ is small, then using the first order Taylor series we can obtain
\begin{equation}\label{left_jacobian_444}
\Gamma_{m}(\phi+\psi)\approx \Gamma_{m}(\phi)Exp(\Gamma_{m+1}(-\phi)\psi)=\Gamma_{m}(\phi)\Gamma_{0}(\Gamma_{m+1}(-\phi)\psi)
\end{equation}
where equation (\ref{Jacobian_left_to_right}) has been used in the above derivation.
The right Jacobian matrix approximation in the BCH formula given in equation (\ref{lie_algebra_perturb_right_Ad}) can also be obtained by equation (\ref{left_jacobian_444}) when $m=0$.

In particular, the auxiliary function for 3-dimensional special orthogonal group introduce by ~\cite{bloesch2013state} is given
\begin{equation}\label{Gamma}
\Gamma_m(\phi)
 \triangleq \sum_{n=0}^{\infty}\frac{1}{(n+m)!}(\phi_{\wedge}^n)
\end{equation}
Then the integrals can be easily expressed and computed by the matrix Taylor series 
\begin{equation}\label{Gamma_0}
\Gamma_0(\phi)=I_3+\frac{\sin||\phi||}{||\phi||}\phi_{\wedge}+\frac{1-\cos||\phi||}{||\phi||^2}\phi_{\wedge}^2=T
\end{equation}
\begin{equation}\label{Gamma_11}
\Gamma_1=I_3+\frac{1-\cos||\phi||}{||\phi||^2}\phi_{\wedge}+\frac{||\phi||-\sin||\phi||}{||\phi||^3}\phi_{\wedge}^2=J=J_l(\phi)
\end{equation}
It is worth noting that $\Gamma_0(\phi)$ is the exponential mapping of $SO(3)$, while $\Gamma_1(\phi)$ is the left Jacobian of $SO(3)$.

Since we have  $\Gamma_2(\phi)\phi_{\wedge}+I_3=\Gamma_1(\phi)$ and $\Gamma_2(\phi)$ can be represented as $\Gamma_2(\phi)=\frac{1}{2!}I_3+x\phi_{\wedge}+y\phi_{\wedge}^2$, then $x$ and $y$ can be obtained by determined coefficient method:
\begin{equation}\label{Gamma_22}
\Gamma_2(\phi)=\frac{1}{2}I_3+\frac{||\phi||-\sin||\phi||}{||\phi||^3}\phi_{\wedge}+\frac{||\phi||^2+2\cos||\phi||-2}{2||\phi||^4}\phi_{\wedge}^2
\end{equation}

Similarly, as we have $\Gamma_3(\phi)\phi_{\wedge}+\frac{1}{2}I_3=\Gamma_2(\phi)$ and $\Gamma_3(\phi)$ can be represented as $\Gamma_3(\phi)=\frac{1}{3!}I_3+a\phi_{\wedge}+b\phi_{\wedge}^2$, then $a$ and $b$ can be obtained by determined coefficient method:
\begin{equation}\label{Gamma_3}
\Gamma_3(\phi)=\frac{1}{3!}I_3+\frac{||\phi||^2+2\cos||\phi||-2}{2||\phi||^4}\phi_{\wedge}+\frac{||\phi||^3-6||\phi||+6\sin||\phi||}{6||\phi||^5}\phi_{\wedge}^2
\end{equation}

It can be verified that 
\begin{equation}\label{left_Jacobian_to_right_Jacobian}
\Gamma_{m}(-\phi)=\Gamma_{m}(\phi)^T
\end{equation}
and
\begin{equation}\label{rotation_for_all_left_and_right}
\Gamma_{m}\left(\Gamma_{0}(\phi)\phi\right)=\Gamma_{0}(\phi)\Gamma_{m}(\phi)\Gamma_{0}(-\phi)\Rightarrow \Gamma_{0}(\phi)\Gamma_{m}(\phi)=\Gamma_{m}(\Gamma_{0}(\phi)\phi)\Gamma_{0}(\phi)
\end{equation}

Equation (\ref{left_Jacobian_to_right_Jacobian}) is true due to the antisymmetric property of the wedge operator in $\mathfrak{so}_3$.

The linearization of a function $\Gamma_{m}(\cdot )$ is the first order Taylor series of the function evaluated as a certain element of the domain. 
If we assume that $\phi$ is small, then using the first order Taylor series we can obtain
\begin{equation}\label{BCH_left}
\Gamma_{m}(\phi+\psi)\approx Exp(\Gamma_{m+1}(\psi)\phi)\Gamma_{m}(\psi)=\Gamma_{0}(\Gamma_{m+1}(\psi)\phi)\Gamma_{m}(\psi)
\end{equation}
It is obvious that the approximate BCH formula that using left Jacobian matrix is the case when $m=0$.

If we assume that $\psi$ is small, then using the first order Taylor series we can obtain
\begin{equation}\label{BCH_1}
\Gamma_{m}(\phi+\psi)\approx \Gamma_{m}(\phi)Exp(\Gamma_{m+1}(-\phi)\psi)=\Gamma_{m}(\phi)\Gamma_{0}(\Gamma_{m+1}(-\phi)\psi)
\end{equation}
It is obvious that the approximate BCH formula that using right Jacobian matrix is the case when $m=0$.

From the definition of Jacobians of $SE_K(3)$ in equation (\ref{left_jacobian_gourp}) and equation (\ref{right_jacobian_gourp}), we get:
\begin{equation}\label{inverse_left_Jacobian_Ad}
\mathcal{J}_l^{-1}=\begin{bmatrix}
J_l^{-1}             & 0          &0      &\cdots & 0 \\
-J_l^{-1}{Q_1}_lJ_l^{-1} &J_l^{-1}    &0      &\cdots & 0\\
-J_l^{-1}{Q_2}_lJ_l^{-1} &0           &J_l^{-1}&\cdots& 0\\
\vdots & \vdots & \vdots &\ddots & 0\\
-J_l^{-1}{Q_K}_lJ_l^{-1} &0&0&\cdots&J_l^{-1}
\end{bmatrix}
\end{equation}
\begin{equation}\label{inverse_right_Jacobian_Ad}
\mathcal{J}_r^{-1}=\begin{bmatrix}
J_r^{-1}             & 0          &0      &\cdots & 0 \\
-J_r^{-1}{Q_1}_rJ_r^{-1} &J_r^{-1}    &0      &\cdots & 0\\
-J_r^{-1}{Q_2}_rJ_r^{-1} &0           &J_r^{-1}&\cdots& 0\\
\vdots & \vdots & \vdots &\ddots & 0\\
-J_r^{-1}{Q_K}_rJ_r^{-1} &0&0&\cdots&J_r^{-1}
\end{bmatrix}
\end{equation}

We have that the singularities of $\mathcal{J}_l$ and $\mathcal{J}_r$ are precisely the same as the singularities of $J_l$ and $J_r$, respectively, since 
\begin{equation}\label{determinant_Jacobian}
\det(\mathcal{J}_l)=(\det(J_l))^{(K+1)},\det(\mathcal{J}_r)=(\det(J_r))^{(K+1)}
\end{equation}
 and having a non-zero determinant is a necessary and sufficient condition for invertibility, and therefore $\mathcal{J}$ has no singularity.

It is worth noting that $\mathcal{J}\mathcal{J}^T>0$ means $\mathcal{J}\mathcal{J}^T$ is positive definite for either the left or right Jacobian matrix.  We can see this through the following factorization:
\begin{equation}\label{Ja cobian_positive_definte}
\begin{aligned}
\mathcal{J}\mathcal{J}^T&=\underbrace{\begin{bmatrix}
I_{3(K+1)} & 0& 0& \cdots & 0\\
Q_1J^{-1}  & I_{3(K+1)}& 0 &\cdots & 0\\
Q_2J^{-1} &0&I_{3(K+1)}&\cdots & 0\\
\vdots & \vdots & \vdots &\ddots & 0\\
Q_KJ^{-1}& 0& 0& \cdots &I_{3(K+1)}
\end{bmatrix}}_{>0}
\underbrace{\begin{bmatrix}
	JJ^T & 0& 0& \cdots & 0\\
	0  & JJ^T& 0 &\cdots & 0\\
	0 &0&JJ^T&\cdots & 0\\
	\vdots & \vdots & \vdots &\ddots & 0\\
	0& 0& 0& \cdots &JJ^T\end{bmatrix}}_{>0}\\
&\quad 
\underbrace{\begin{bmatrix}
I_{3(K+1)} & J^{-T}Q_1^T& J^{-T}Q_2^T& \cdots & J^{-T}Q_K^T\\
0  & I_{3(K+1)}& 0 &\cdots & 0\\
0 &0&I_{3(K+1)}&\cdots & 0\\
\vdots & \vdots & \vdots &\ddots & 0\\
0& 0& 0& \cdots &I_{3(K+1)}\end{bmatrix}}_{>0}>0
\end{aligned}
\end{equation}
where we have used $JJ^T>0$ which has been proved in \cite{barfoot2017state}.
\subsection{Distance, Volume, Integration}
We need to think about the concepts of distance, volume, and integration differently for Lie groups than for vectorspaces. 

There are two common ways to define the difference of two elements of the matrix Lie group $SE_K(3)$ and $Ad(SE_K(3))$, and they are called the right and left distance metrics, respectively:
\begin{equation}\label{left_distance_metrics}
\xi_{12}=\mathcal{L}^{-1}(\log_G(T_1^{-1}T_2))=\mathfrak{L}^{-1}(\log_G(\mathcal{T}_1^{-1}\mathcal{T}_2)),\quad \text{left invariant}
\end{equation}
\begin{equation}\label{right_distance_metrics}
\xi_{21}=\mathcal{L}^{-1}(\log_G(T_2T_1^{-1}))=\mathfrak{L}^{-1}(\log_G(\mathcal{T}_2\mathcal{T}_1^{-1})),\quad \text{right invariant}
\end{equation}

Then we can define the $(K+3)\times (K+3)$ and $3(K+1)\times 3(K+1)$ inner products for $\mathfrak{se}_k(3)$ as
\begin{equation}\label{K_3_inner_product}
<\mathcal{L}(\xi_1),\mathcal{L}(\xi_2)>=\xi_1^T\xi_2
\end{equation}
\begin{equation}\label{3_K_1_inner_product}
<\mathfrak{L}(\xi_1),\mathfrak{L}(\xi_2)>=\xi_1^T\xi_2
\end{equation}

Suppose there are weight matrices that make the inner product completely defined, the weight matrices should be diagonal:
\begin{equation}\label{K_3_inner_product_weight}
\begin{aligned}
&\quad <\mathcal{L}(\xi_1),\mathcal{L}(\xi_2)>=tr\left( \mathcal{L}(\xi_1)   \begin{bmatrix}
cI_3 & 0       & 0      & \cdots & 0\\
\bf{0}^T    & d_1 & 0      & \cdots & 0\\
\bf{0}^T    & 0       & d_2 & \cdots & 0\\
\vdots&\vdots  &\vdots  &\ddots  & \vdots\\
\bf{0}^T&    0&    0& \cdots &d_K
\end{bmatrix}\mathcal{L}(\xi_2) \right)\\
&=-c\cdot tr({\theta_1}_{\wedge}{\theta_2}_{\wedge})+ tr(\sum_{k=1}^{K}d_k({t_1}_k)({t_2}_k)^T=\xi_1^T\xi_2
\end{aligned}
\end{equation}
\begin{equation}\label{3_K_1_inner_product_weight}
\begin{aligned}
&\quad <\mathfrak{L}(\xi_1),\mathfrak{L}(\xi_2)>=tr\left( \mathfrak{L}(\xi_1)   \begin{bmatrix}
aI_3 & 0       & 0      & \cdots & 0\\
0    & b_1 I_3 & 0      & \cdots & 0\\
0    & 0       & b_2I_3 & \cdots & 0\\
\vdots&\vdots  &\vdots  &\ddots  & \vdots\\
0&    0&    0& \cdots &b_KI_3
\end{bmatrix}\mathfrak{L}(\xi_2) \right)\\
&=-(a+b_1+b_2+\cdots+b_K)tr({\theta_1}_{\wedge}{\theta_2}_{\wedge})-a\cdot tr(\sum_{k=1}^{K}({t_1}_k)_{\wedge}({t_2}_k)_{\wedge})=\xi_1^T\xi_2
\end{aligned}
\end{equation}
where $tr(\cdot)$ means the trace of matrix and the unknown coefficients $c, d_1, d_2, \cdots, d_K$ and $a, b_1, b_2, \cdots, b_K$ can be solved such that 
\begin{equation}\label{coefficient_metric_weight}
c=\frac{1}{2},d_1=d_2=\cdots=d_K=1, 
a=\frac{1}{2},b_1=b_2=\cdots=b_K=0
\end{equation}

The right and left distance are
\begin{equation}\label{right_distance_definition_2}
\xi_{12}=\sqrt{<\mathcal{L}(\xi_{12}),\mathcal{L}(\xi_{12})>}=\sqrt{<\mathfrak{L}(\xi_{12}),\mathfrak{L}(\xi_{12})>}=\sqrt{\xi_{12}^T\xi_{12}}=||\xi_{12}||
\end{equation}
\begin{equation}\label{left_distance_definition_2}
\xi_{21}=\sqrt{<\mathcal{L}(\xi_{21}),\mathcal{L}(\xi_{21})>}=\sqrt{<\mathfrak{L}(\xi_{21}),\mathfrak{L}(\xi_{21})>}=\sqrt{\xi_{21}^T\xi_{21}}=||\xi_{21}||
\end{equation}

Using the parameterization
\begin{equation}\label{parameterization_T}
T=\exp_G(\mathcal{L}(\xi))
\end{equation}
and the perturbation
\begin{equation}\label{perturbation_T}
T'=\exp_G(\mathcal{L}(\xi+\delta\xi))
\end{equation}
by combining with equation (\ref{lie_algebra_perturb_left}) and equation (\ref{lie_algebra_perturb_right}), the differences relative to $T$ are
\begin{equation}\label{left_difference}
\begin{aligned}
\mathcal{L}^{-1}(\log_G(\delta T_r))&=\mathcal{L}^{-1}(\log_G(T^{-1} T'))=\mathcal{L}^{-1}(\log_G(\exp_G(\mathcal{L}(\xi))^{-1} \exp_G(\mathcal{L}(\xi+\delta\xi))))\\
&=\mathcal{L}^{-1}(\log_G(\exp_G(\mathcal{L}(-\xi)) \exp_G(\mathcal{L}(\xi)+\mathcal{L}(\delta\xi))))\approx \mathcal{J}_r\delta\xi
\end{aligned}
\end{equation}
\begin{equation}\label{right_difference}
\begin{aligned}
\mathcal{L}^{-1}(\log_G(\delta T_l))&=\mathcal{L}^{-1}(\log_G( T'T^{-1}))=\mathcal{L}^{-1}(\log_G(\exp_G(\mathcal{L}(\xi+\delta\xi)\exp_G(\mathcal{L}(\xi))^{-1})))\\
&=\mathcal{L}^{-1}(\log_G(\exp_G(\mathcal{L}(\delta\xi)+\mathcal{L}(\xi)) \exp_G(\mathcal{L}(-\xi)) ))\approx \mathcal{J}_l\delta\xi
\end{aligned}
\end{equation}

The right and left infinitesimal volume elements are
\begin{equation}\label{right_infinitesimal}
dT_r=|\det(\mathcal{J}_r)|d\xi
\end{equation}
\begin{equation}\label{left_infinitesimal}
dT_l=|\det(\mathcal{J}_l)|d\xi
\end{equation}

We have that 
\begin{equation}\label{determinant_left_right}
\det(\mathcal{J}_l)=\det(\mathcal{T}\mathcal{J}_r)=\det(\mathcal{T})\det(\mathcal{J}_r)=\det(\mathcal{J}_r)
\end{equation}
since $\det(\mathcal{T})=(\det(R))^{K+1}=1$. We can therefore write
\begin{equation}\label{infinitesimal}
dT=|\det(\mathcal{J})|d\xi
\end{equation}
for our integration volume. Finally, we have that
\begin{equation}\label{determinant_J}
|\det(\mathcal{J})|=|\det(J)|^{K+1}=\left(2\frac{1-\cos\theta}{\theta^2}\right)^{K+1}=\left(\frac{\sin(\frac{\theta}{2})}{\frac{\theta}{2}}\right)^{2(K+1)}
\end{equation}

To integrate functions over $SE_K(3)$, we can now use our infinitesimal volume in the calculation:
\begin{equation}\label{integration_SE_K_3}
\int_{SE_K(3)}f(T)dT=\int_{|\theta|<\pi,\mathbb{R}^{3K}}f(\xi)|\det(\mathcal{J})|d\xi
\end{equation}
where we limit $\theta$ to the ball of radius $\pi$ (due to the surjective-only nature of the exponential map) but let $t_k\in\mathbb{R}^3,k=1,\cdots,K$.
\subsection{Interpolation}
We let $T=\exp_G(\mathcal{L}(\xi))$, $T_1=\exp_G(\mathcal{L}(\xi_1))$, $T_2=\exp_G(\mathcal{L}(\xi_2))\in SE_K(3)$ with $\xi, \xi_1, \xi_2\in \mathbb{R}^{3(K+1)}$. If we are able to make the assumption that $\xi_{21}$ is small in the sense of distance from equation (\ref{right_distance_metrics}), then we have  
\begin{equation}\label{interpolation}
\begin{aligned}
\xi&=\mathcal{L}^{-1}(\log_G(T))=\mathcal{L}^{-1}(\log_G((T_2T_1^{-1})^{\alpha}T_1))=\mathcal{L}^{-1}(\log_G(\exp_G(\alpha\mathcal{L}(\xi_{21}))\exp_G(\mathcal{L}(\xi_1))))\\
&\approx\alpha J_l(\xi_1)^{-1}\xi_{21}+\xi_1
\end{aligned}
\end{equation}
which is a form of linear interpolation.

Another case worth noting when $T_1=I_{K+3}$, whereupon
\begin{equation}\label{interpolation_T_1}
T=T_2^{\alpha}, \xi=\alpha\xi_2
\end{equation}
with no approximation.
\subsection{Homogeneous Points}
Points in $\mathbb{R}^3$ can be represented using $4\times 1$ homogeneous coordinates as follows
\begin{equation}\label{honogeneous_coordinates}
p=\begin{bmatrix}
sx\\sy\\sz\\s
\end{bmatrix}=\begin{bmatrix}
\varepsilon\\\eta
\end{bmatrix}
\end{equation}
where s is some real, nonzero scalar, $\varepsilon\in\mathbb{R}^3$, and $\eta$ is scalar.
When s is zero, it is not possible to convert back to $\mathbb{R}^3$, as this case represents points that are infinitely far away~\cite{barfoot2017state}.

It is useful to define the operator
\begin{equation}\label{operator_o}
p^{\odot}=\begin{bmatrix}
\varepsilon\\\eta
\end{bmatrix}^{\odot}=\begin{bmatrix}
-\varepsilon_{\wedge} & \eta I_3\\0_{1\times 3}&0_{1\times 3}
\end{bmatrix}_{4\times 6},\varepsilon\in\mathbb{R}^3,\eta\in\mathbb{R}
\end{equation}
such that $\mathcal{L}(x)p=p^{\odot}x$ holds, where $x\in\mathbb{R}^6,p\in\mathbb{R}^4$.

It is also useful to define the operator
\begin{equation}\label{operator_extended}
\forall p=\begin{bmatrix}
\varepsilon\\\eta_1\\\eta_2\\\vdots\\\eta_K
\end{bmatrix}
\in\mathbb{R}^{K+3},p^{\odot}=\begin{bmatrix}
\varepsilon\\\eta_1\\\eta_2\\\vdots\\\eta_K
\end{bmatrix}^{\odot}=\begin{bmatrix}
-\varepsilon_{\wedge} &\eta_1I_3&\eta_2I_3&\cdots & \eta_KI_3\\
0_{K\times 3}&0_{K\times 3}&0_{K\times 3}&\cdots&0_{K\times 3}
\end{bmatrix}_{(K+3)\times 3(K+1)}
\end{equation}
where $\varepsilon\in\mathbb{R}^3$ and $\eta_1,\eta_2,\cdots,\eta_K\in\mathbb{R}$, such that $\forall x\in\mathbb{R}^{3(K+1)},\mathcal{L}(x)p=p^{\odot}x$ holds.

We also have the following identities
\begin{equation}\label{community_operator_o}
\forall T\in SE_K(3), (Tp)^{\odot}=Tp^{\odot}\mathcal{T}^{-1}
\end{equation}
\begin{equation}\label{community_operator_o_1}
\forall T\in SE_K(3), {(Tp)^{\odot}}^T(Tp)^{\odot}=\mathcal{T}^{-T}{p^{\odot}}^Tp^{\odot}\mathcal{T}^{-1}
\end{equation}
\subsection{Calculus and Optimization}
On occasion, we construct  functions related to the Lie group elements and then discuss the perturbation of the elements. Consequently, we face the problem taking derivatives with respect to the Lie algebras so as to find the minimal perturbation. There two ways of thinking about it:
\begin{itemize}
	\item One way is to represent the Lie group with the Lie algebra, and then take the derivative of the Lie algebra according to the Euclidean space addition.
	\item Another way is to multiply the Lie group left or right by a small perturbation, and then take the derivative of that perturbation. 
\end{itemize}

For the group $Ad(SE_K(3))$, we may want to take the derivative of the product of a $3(K+1)\times 3(K+1)$ transformation matrix and a $3(K+1)\times 1$ column, with respect to the $3(K+1)\times 1$ pose variable. Assume $
\mathcal{T}\in \mathbb{R}^{3(K+1)\times 3(K+1)}$, $x\in \mathbb{R}^{3(K+1)}$ and $\xi\in \mathbb{R}^{3(K+1)}$, taking the derivative of the group left action of $Ad(SE_K(3))$ on vector $x$ along $\xi$, we have
\begin{equation}\label{jacobian_wrt_algebra_SE_K}
\begin{aligned}
&\quad \frac{\partial(\mathcal{T}x)}{\partial \xi}=\frac{\partial(\exp_G(\mathfrak{L}(\xi))x)}{\partial \xi}=\lim\limits_{\delta\xi\rightarrow \bf{0}} \frac{\exp_G(\mathfrak{L}(\xi+\delta\xi))x-\exp_G(\mathfrak{L}(\xi))x}{ \delta \xi}\\
&=\lim_{\delta\xi\rightarrow \bf{0}}\frac{\exp_G(\mathfrak{L}(\mathcal{J}_l\delta\xi))\exp_G(\mathfrak{L}(\xi))x-\exp_G(\mathfrak{L}(\xi))x}{\delta \xi}\\
&=\lim_{\delta\xi\rightarrow \bf{0}}\frac{(I_{3(K+1)}+\mathfrak{L}(\mathcal{J}_l\delta\xi))\exp_G(\mathfrak{L}(\xi))x-\exp_G(\mathfrak{L}(\xi))x}{\delta \xi}=\lim_{\delta\xi\rightarrow \bf{0}}\frac{\mathfrak{L}(\mathcal{J}_l\delta\xi)\exp_G(\mathfrak{L}(\xi))x}{\delta \xi}
\end{aligned}
\end{equation}

Using the property stated in the equation (\ref{antisymetry}), we have
\begin{equation}\label{jacobian_wrt_algebra_SE_K_final}
\frac{\partial(\mathcal{T}x)}{\partial \xi}=\lim_{\delta\xi\rightarrow \bf{0}}\frac{-\mathfrak{L}(\exp_G(\mathfrak{L}(\xi))x)\mathcal{J}_l\delta \xi}{\delta\xi}=-\mathfrak{L}(\mathcal{T}x)\mathcal{J}_l
\end{equation}

However, since the result contain a complex form of the left Jacobian matrix $\mathcal{J}_l$, we do not want to calculate it. 
Another way to perturb $T$ by left multiplying a perturbation $\delta \mathcal{T}=\exp_G(\mathfrak{L}(\delta \xi))$ as:
\begin{equation}\label{jacobian_wrt_algebra_Ad_SE_K}
\begin{aligned}
&\quad \frac{\partial(\mathcal{T}x)}{\partial \delta\xi}=\frac{\partial(\exp_G(\mathfrak{L}(\xi))x)}{\partial \delta\xi}=\lim\limits_{\delta\xi\rightarrow \bf{0}} \frac{\delta\mathcal{T}\exp_G(\mathfrak{L}(\xi))x-\exp_G(\mathfrak{L}(\xi))x}{ \delta \xi}\\
&=\lim_{\delta\xi\rightarrow \bf{0}}\frac{\exp_G(\mathfrak{L}(\delta\xi))\exp_G(\mathfrak{L}(\xi))x-\exp_G(\mathfrak{L}(\xi))x}{\delta \xi}\\
&=\lim_{\delta\xi\rightarrow \bf{0}}\frac{(I_{3(K+1)}+\mathfrak{L}(\delta\xi))\exp_G(\mathfrak{L}(\xi))x-\exp_G(\mathfrak{L}(\xi))x}{\delta \xi}=\lim_{\delta\xi\rightarrow \bf{0}}\frac{\mathfrak{L}(\delta\xi)\mathcal{T}x}{\delta \xi}=-\mathfrak{L}(\mathcal{T}x)
\end{aligned}
\end{equation}

Meanwhile, the targets are three-dimensional points in space when we usually want to manipulate. We turn our attention to the homogeneous points in space.
As the Lie group $SE_K(3)$ has $K$ translational components, we define a new matrix that combines $K$ homogeneous points as follows:
\begin{equation}\label{P_K_homogeneous_points}
Q=\begin{bmatrix}
q_1&q_2&q_3&\cdots&q_K\\
1  &0  &0  &\cdots&0\\
0  &1  &0  &\cdots&0\\
\vdots&\vdots&\vdots&\ddots&\vdots\\
0  &0  &0  &\cdots&1
\end{bmatrix}\in \mathbb{R}^{(K+3)\times K},q_k\in \mathbb{R}^3,k=1,\cdots,K
\end{equation}

It is easy to verify that
\begin{equation}\label{T_p_m}
\begin{aligned}
&(Tq_m)^{\odot}=\left[\begin{array}{c}
\begin{array}{cccc}
-\mathcal{L}{(Rq_m+p_m)} &\bf{0}_{3\times(m-1)} &I_{3}     &\bf{0}_{3\times(K-m)}
\end{array}\\
\hdashline[2pt/2pt]
O_{K\times 3(K+1)}
\end{array} \right] \\
=&\begin{bmatrix}
E_m\\O_{K\times 3(K+1)}
\end{bmatrix}
\in \mathbb{R}^{(K+3)\times 3(K+1)},m=1,2,\cdots,K
\end{aligned}
\end{equation}

Analogously, The Jacobian of $K$ transformed point with respect to the Lie algebra vector representing the transformation is
\begin{equation}\label{jacobian_wrt_algebra}
\begin{aligned}
&\quad \frac{\partial(TQ)}{\partial \xi}=\frac{\partial(\exp_G(\mathcal{L}(\xi))Q)}{\partial \xi}=\lim_{\delta\xi\rightarrow \bf{0}}\frac{\exp_G(\mathcal{L}(\xi+\delta\xi))Q-\exp_G(\mathcal{L}(\xi))Q}{\delta\xi}\\
&=\lim_{\delta\xi\rightarrow \bf{0}}\frac{\exp_G(\mathcal{L}(\mathcal{J}_l\delta\xi))\exp_G(\mathcal{L}(\xi))Q-\exp_G(\mathcal{L}(\xi))Q}{\delta \xi}\\
&=\lim_{\delta\xi\rightarrow \bf{0}}\frac{(I_{3(K+1)}+\mathcal{L}(\mathcal{J}_l\delta\xi))\exp_G(\mathcal{L}(\xi))Q-\exp_G(\mathcal{L}(\xi))Q}{\delta \xi}=\lim_{\delta\xi\rightarrow \bf{0}}\frac{\mathcal{L}(\mathcal{J}_l\delta\xi)\exp_G(\mathcal{L}(\xi))Q}{\delta \xi}\\
&=\lim_{\delta\xi\rightarrow \bf{0}}\frac{\mathcal{L}\left(\begin{bmatrix}
	J_l\delta \phi_{\wedge}\\ Q_1\delta\phi+J_l\delta t_1 \\Q_2\delta\phi+J_l\delta t_2 \\\vdots \\Q_K\delta\phi+J_l\delta t_K
	\end{bmatrix} \right)\begin{bmatrix}
	Rq_1+p_1 &Rq_2+p_2 &Rq_3+p_3& \cdots &Rq_K+t_p\\
	1        &   0     & 0      &\cdots  & 0\\
	0        &   1     &0       &\cdots  & 0\\
	0&0&1&\cdots  & 0\\
	0&0&0&\cdots  & 1
	\end{bmatrix}}{[\delta\phi,\delta t_1,\cdots,\delta t_K]^T}\\
&=\lim_{\delta\xi\rightarrow \bf{0}}\frac{\left[\begin{array}{c}
	\begin{array}{cccc}
	(J_l\delta\phi)_{\wedge}(Rq_1+p_1)+Q_1\delta\phi+J_l\delta t_1&\cdots& (J_l\delta\phi)_{\wedge}(Rq_K+p_K)+Q_K\delta\phi+J_l\delta t_K
	\end{array}\\
	\hdashline[2pt/2pt]
	O_{K\times K}
	\end{array} \right]}{[\delta\phi,\delta t_1,\cdots,\delta t_K]^T}\\
&=\left[\begin{array}{c}
\begin{array}{cccc}
E_1\mathcal{J}_l&E_2\mathcal{J}_l&\cdots& E_K\mathcal{J}_l
\end{array}\\
\hdashline[2pt/2pt]
O_{K\times 3K(K+1)}
\end{array} \right] 
\in \mathbb{R}^{(K+3)\times 3K(K+1)}
\end{aligned}
\end{equation}
where $E_m=\begin{bmatrix}
-\mathcal{L}{(Rq_m+p_m)} &\bf{0}_{3\times(m-1)} &I_{3}     &\bf{0}_{3\times(K-m)}
\end{bmatrix}\in \mathbb{R}^{3\times 3(K+1)},m=1,\cdots,K$ and $\mathcal{J}_l$ is  the left Jacobian of $SE_K(3)$. 

Using the equation (\ref{T_p_m}), the above equation can be denoted as
\begin{equation}\label{jacobian_wrt_algebra_operator}
\quad \frac{\partial(TQ)}{\partial \xi}=\begin{bmatrix}
(Tq_1)^{\odot}\mathcal{J}_l& (Tq_2)^{\odot}\mathcal{J}_l& \cdots & (Tq_K)^{\odot}\mathcal{J}_l
\end{bmatrix}
\end{equation}

Again, the result still contains a complex form of the left Jacobian matrix $\mathcal{J}_l$, we do not want to calculate it. 
Another way to perturb $T$ by left multiplying a perturbation $\delta T=\exp_G(\mathcal{L}(\delta \xi))$ as:
\begin{equation}\label{jacobian_left_multiplication}
\begin{aligned}
&\quad\frac{\partial(TQ)}{\partial \delta\xi}=\frac{\partial(\exp_G(\mathcal{L}(\xi))Q)}{\partial \delta \xi}=\lim_{\delta\xi\rightarrow \bf{0}}\frac{\exp_G(\mathcal{L}(\delta\xi))\exp_G(\mathcal{L}(\xi))Q-\exp_G(\mathcal{L}(\xi))Q}{\partial \xi}\\
&=\lim_{\delta\xi\rightarrow \bf{0}}\frac{(I_{3(K+1)}+\mathcal{L}(\delta\xi))\exp_G(\mathcal{L}(\xi))Q-\exp_G(\mathcal{L}(\xi))Q}{\partial \xi}=\lim_{\delta\xi\rightarrow \bf{0}}\frac{\mathcal{L}(\delta\xi)\exp_G(\mathcal{L}(\xi))Q}{\partial \xi}\\
&=\lim_{\delta\xi\rightarrow \bf{0}}\frac{\begin{bmatrix}
	\delta \phi_{\wedge} & \delta t_1 &\delta t_2 &\cdots &\delta t_K\\
	\bf{0}^T &0 &0 &\cdots & 0\\
	\bf{0}^T &0 &0 &\cdots & 0\\
	\vdots & \vdots & \vdots & \ddots&\vdots\\
	\bf{0}^T &0 &0 &\cdots & 0
	\end{bmatrix}\begin{bmatrix}
	Rq_1+p_1 &Rq_2+p_2 &Rp_3+t_3& \cdots &Rq_K+p_K\\
	1        &   0     & 0      &\cdots  & 0\\
	0        &   1     &0       &\cdots  & 0\\
	0&0&1&\cdots  & 0\\
	0&0&0&\cdots  & 1
	\end{bmatrix}}{[\delta\phi,\delta t_1,\cdots,\delta t_K]^T}\\
&=\lim_{\delta\xi\rightarrow \bf{0}}\frac{\left[\begin{array}{c}
	\begin{array}{cccc}
	\phi_{\wedge}(Rq_1+p_1)+\delta t_1&\phi_{\wedge}(Rq_2+p_2)+\delta t_2&\cdots& \phi_{\wedge}(Rq_K+p_K)+\delta t_K
	\end{array}\\
	\hdashline[2pt/2pt]
	O_{K\times K}
	\end{array} \right]}{[\delta\phi,\delta t_1,\cdots,\delta t_K]^T}\\
&=\left[\begin{array}{c}
\begin{array}{cccc}
E_1&E_2&\cdots& E_K
\end{array}\\
\hdashline[2pt/2pt]
O_{K\times 3K(K+1)}
\end{array} \right] 
\in \mathbb{R}^{(K+3)\times 3K(K+1)}
\end{aligned}
\end{equation}
where $E_m=\begin{bmatrix}
-\mathcal{L}{(Rq_m+p_m)} &\bf{0}_{3\times(m-1)} &I_{3}     &\bf{0}_{3\times(K-m)}
\end{bmatrix}\in \mathbb{R}^{3\times 3(K+1)},m=1,\cdots,K$. It is worth noting that we will not calculate Jacobian matrix $\mathcal{J}_l$ anymore, which makes the perturbation more practical and useful.

Similarly, using the equation (\ref{T_p_m}), the above equation can be denoted as
\begin{equation}\label{jacobian_wrt_left_mulplication_operator}
\quad \frac{\partial(TQ)}{\partial \delta \xi}=\begin{bmatrix}
(Tq_1)^{\odot}& (Tq_2)^{\odot}& \cdots & (Tq_K)^{\odot}
\end{bmatrix}
\end{equation}

Finally, for optimization, suppose we have a general nonlinear, quadratic cost function of a transformation of the form
\begin{equation}\label{cost_function_matrix}
C(T)=\frac{1}{2}\sum_{n}(u_n(TQ))^2
\end{equation}
where $u_n(\cdot)$ are nonlinear functions of matrices and $Q\in \mathbb{R}^{(K+3)\times K}$ are combined by K three-dimensional points expressed in homogeneous coordinates. This could be solved by the Gauss-Newton algorithm.
\section{Kinematics of Lie group $SE_K(3)$}
We have seen how the geometry of a Lie group works. The next step
is to allow the geometry to change over time. We will work out the
kinematics associated with our matrix Lie group $SE_K(3)$.
\subsection{Lie Group and Lie Algebra}
Let us rewrite the exponential map from the Lie algebra to the matrix Lie group $SE_K(3)$ here:
\begin{equation}\label{exponential_map_transformation}
\begin{aligned}
T&=\exp_G(\mathcal{L}(\xi))=\begin{bmatrix}
R&p_1&p_2& \cdots & p_K\\
\bf{0}^T&1&0&\cdots &0\\
\bf{0}^T&0&1&\cdots&0\\
\vdots&\vdots&\vdots&\ddots&\vdots\\
\bf{0}^T&0&0&\cdots&1
\end{bmatrix}, R\in SO(3), p_k\in \mathbb{R}^3, k=1,\cdots,K
\end{aligned}
\end{equation}

Suppose the kinematics in terms of separated translation and rotation are given by
\begin{equation}\label{kinematics}
\begin{aligned}
\dot{p}_k&={\omega_l}_{\wedge}p_k+{\nu_l}_k=R{\nu_r}_k, k=1,\cdots,K\\
\dot{R}&={\omega_l}_{\wedge}R=R{\omega_r}_{\wedge}
\end{aligned}
\end{equation}
where $\omega_l, {\nu_l}_k, k=1, \cdots, K$ are the rotational and translational left velocities, respectively; $\omega_r, {\nu_r}_k, k=1, \cdots, K$ are the rotational and translational right velocities, respectively.

Using the transformation matrices, the kinematic differential equation in terms of $T$ can be written equivalently as 
\begin{equation}\label{equivalently_kinematic}
\begin{aligned}
&\dot{T}=\mathcal{L}(\overline{\omega}_l)T=T\mathcal{L}(\overline{\omega}_r)=T\mathcal{L}(\overline{\omega}_r)T^{-1}T=\mathcal{L}({Ad_T}\overline{\omega}_r)T \\
\Rightarrow &\mathcal{L}(\overline{\omega}_l)=\dot{T}T^{-1}\quad \mathcal{L}(\overline{\omega}_r)=T^{-1}\dot{T} \quad \overline{\omega}_l={Ad_T}\overline{\omega}_r
\end{aligned}
\end{equation}
where 
\begin{equation}\label{overline_omega}
\overline{\omega}_l=\begin{bmatrix}
\omega_l\\ \nu_{l1}\\ \nu_{l2}\\ \vdots\\ \nu_{lK}
\end{bmatrix}\in \mathbb{R}^{3(K+1)},\overline{\omega}_r=\begin{bmatrix}
\omega_r\\ \nu_{r1}\\ \nu_{r2}\\ \vdots\\ \nu_{rK}
\end{bmatrix}\in \mathbb{R}^{3(K+1)},\begin{bmatrix}
\omega_l=R\omega_r\\
\nu_{l1}={p_1}_{\wedge}R\omega_r+R\nu_{r1}\\ 
\nu_{l2}={p_2}_{\wedge}R\omega_r+R\nu_{r2}\\ 
\vdots\\ 
\nu_{lK}={p_K}_{\wedge}R\omega_r+R\nu_{rK}
\end{bmatrix}
\end{equation}
$\overline{\omega}_l$ and $\overline{\omega}_r$ are the generalized left velocity and the generalized right velocity, respectively. It is worth noting that these equations are singularity-free but still have the constraint that $RR^{-1}=I_3$.

The kinematics can also be written using the adjoint quantities:
\begin{equation}\label{kinematics_adjoint}
\dot{\mathcal{T}}=\mathfrak{L}(\overline{\omega}_l)\mathcal{T}=\mathcal{T}\mathfrak{L}(\overline{\omega}_r)
\end{equation}

To see the equivalent kinematics in terms of the Lie algebra, we need to differentiate the group element $T \in SE_K(3)$ by analogy to the groups $SO(3)$ and $SE(3)$, which comes from the general expression for the time derivative of the matrix exponential~\cite{barfoot2017state}:
\begin{equation}\label{time_derivative_T}
\begin{aligned}
\dot{T}=\frac{d}{dt}\exp_G(\mathcal{L}(\xi))&=\int_{0}^{1}\exp_G(\alpha\mathcal{L}(\xi))\mathcal{L}(\dot{\xi})\exp_G((1-\alpha)\mathcal{L}(\xi))d\alpha\\
&=\int_{0}^{1}\exp_G((1-\alpha)\mathcal{L}(\xi))\mathcal{L}(\dot{\xi})\exp_G(\alpha\mathcal{L}(\xi))d\alpha
\end{aligned}
\end{equation}
or equivalently,
\begin{equation}\label{time_derivative_T_1}
\dot{T}T^{-1}=\int_{0}^{1}T^{\alpha}\mathcal{L}(\dot{\xi})T^{-\alpha}d\alpha=\int_{0}^{1}\mathcal{L}(\mathcal{T}^{\alpha}\dot{\xi})d\alpha=\mathcal{L}((\int_{0}^{1}\mathcal{T}^{\alpha}d\alpha)\dot{\xi})=\mathcal{L}(\mathcal{J}_l\dot{\xi})
\end{equation}
\begin{equation}\label{time_derivative_T_r}
T^{-1}\dot{T}=\int_{0}^{1}T^{-\alpha}\mathcal{L}(\dot{\xi})T^{\alpha}d\alpha=\int_{0}^{1}\mathcal{L}(\mathcal{T}^{-\alpha}\dot{\xi})d\alpha=\mathcal{L}((\int_{0}^{1}\mathcal{T}^{-\alpha}d\alpha)\dot{\xi})=\mathcal{L}(\mathcal{J}_r\dot{\xi})
\end{equation}
where $\mathcal{J}_l=\int_{0}^{1}\mathcal{T}^{\alpha}d\alpha$ is the left Jacobian for $SE_K(3)$; $\mathcal{J}_r=\int_{0}^{1}\mathcal{T}^{-\alpha}d\alpha$ is the right Jacobian for $SE_K(3)$. Comparing equation (\ref{equivalently_kinematic}) and equation (\ref{time_derivative_T_1}), we can get the relationship between generalized velocity and the pose parameter derivative:
\begin{equation}\label{Lie_algebra_kinematic}
\overline{\omega}_l=\mathcal{J}_l\dot{\xi},\overline{\omega}_r=\mathcal{J}_r\dot{\xi}
\end{equation}
or 
\begin{equation}\label{Lie_algebra_kinematic_1}
\dot{\xi}=\mathcal{J}_l^{-1}\overline{\omega}_l=(Ad_T\mathcal{J}_r)^{-1}(Ad_T\overline{\omega}_l)=\mathcal{J}_r^{-1}\overline{\omega}_r
\end{equation}
for our equivalent kinematics in terms of the Lie algebra. Again, these equations are now free of constraints.

Then we can derive a $SE_K(3)$ Jacobian identity similar to the $SO(3)$ group and $SE(3)$ group\cite{barfoot2017state}:
\begin{equation}\label{Jacobian_Identity}
\dot{\mathcal{J}}(\xi)-\mathfrak{L}(\overline{\omega})\mathcal{J}(\xi) \equiv \frac{\partial \overline{\omega}}{\partial \xi}
\end{equation}

Starting from the right-hand side, we get
\begin{equation}\label{Proof_Jacobian_Identity}
\begin{aligned}
\frac{\partial \overline{\omega}}{\partial \xi}&=\frac{\partial }{\partial \xi}(\mathcal{J}(\xi)\dot{\xi})=\frac{\partial }{\partial \xi}\left(\underbrace{\int_{0}^{1}\mathcal{T}(\xi)^{\alpha}d\alpha}_{\mathcal{J}(\xi)}\dot{\xi}\right)=\int_{0}^{1}\frac{\partial }{\partial \xi}\left(\mathcal{T}(\alpha\xi)\dot{\xi}\right)d\alpha\\
&=-\int_{0}^{1}\mathfrak{L}\left(\mathcal{T}(\alpha\xi)\dot{\xi}\right)\alpha\mathcal{J}(\alpha\xi)d\alpha
\end{aligned}
\end{equation}

Noting that
\begin{equation}\label{Jacobian_Transformation}
\frac{d}{d\alpha}\left(\alpha\mathcal{J}(\alpha\xi)\right)=\mathcal{T}(\alpha\xi), \int\mathcal{T}(\alpha\xi)d\alpha=\alpha\mathcal{J}(\alpha\xi)
\end{equation}
we can then integrate by parts to see that
\begin{equation}\label{Proof_Jacobian_Identity_final}
\begin{aligned}
\frac{\partial \overline{\omega}}{\partial \xi}&=-\underbrace{\mathfrak{L}\left(\alpha \mathcal{J}(\alpha\xi)\dot{\xi}\right)\alpha\mathcal{J}(\alpha\xi)|_{\alpha=0}^{\alpha=1}}_{\mathfrak{L}(\overline{\omega})\mathcal{J}(\xi)}+\int_{0}^{1}\underbrace{\mathfrak{L}(\alpha\mathcal{J}_l(\alpha\xi)\dot{\xi})\mathcal{T}(\alpha\xi)}_{\dot{\mathcal{T}}(\alpha\xi)}d\alpha\\
&=-\mathfrak{L}(\overline{\omega})\mathcal{J}(\xi)+\frac{d}{dt}\underbrace{\int_{0}^{1}\mathcal{T}(\xi)^{\alpha}d\alpha}_{\mathcal{J}(\xi)}=\dot{\mathcal{J}}(\xi)-\mathfrak{L}(\overline{\omega})\mathcal{J}(\xi)
\end{aligned}
\end{equation}
which is the desired result.
\subsection{Linearization in Lie Group and Lie Algebra}
We can also perturb our kinematic about some nominal solution, both in the Lie group and the Lie algebra. We begin with the Lie group $SE_K(3)$. Consider the following perturbed matrix $T'\in SE_K(3)$:
\begin{equation}\label{perturbation_kinematic}
T'=\exp_G(\mathcal{L}(\delta\xi))T\approx (I_{K+3}+\mathcal{L}(\delta \xi))T
\end{equation}
where $T\in SE_K(3)$ and $\xi\in \mathbb{R}^{3(K+1)}$ is a perturbation. The perturbed kinematics,
  \begin{equation}\label{perturbed_kinematic}
  \dot{T}'=\mathcal{L}(\overline{\omega}_l')T'
  \end{equation}
  can then be broken into nominal and perturbation kinematics:
  \begin{equation}
  \begin{aligned}
  \text{nominal kinematics}: & \dot{T}=\mathcal{L}(\overline{\omega}_l)T\\
  \text{perturbation kinematics}:& \delta \dot{\xi}=\mathfrak{L}(\overline{\omega}_l)\delta\xi+\delta\overline{\omega}_l
  \end{aligned}
  \end{equation}
  where $\overline{\omega}_l'=\overline{\omega}_l+\delta\overline{\omega}_l$. These can be integrated separately and combined to provide the complete solution approximately.
\section{Uncertainties on Matrix Lie Group $SE_K(3)$}
Aiming at defining random variables on matrix Lie group $SE_K(3)$, we can not apply the usual method of additive Gaussian noise for $T_1,T_2\in SE_K(3)$ as $SE_K(3)$ is not a vector space, i.e., generally $T_1+T_2\notin SE_K(3)$ does not hold.
In contract, we adopt the concentrated Gaussian distribution\cite{barfoot2014associating} using differential geometry tools to define a Gaussian distribution directly on the manifold.
Specifically, in matrix Lie group $SE_K(3)$, there are both left multiplication and right multiplication depending on whether the noise is multiplied through the left or right of the group element:
\begin{equation}\label{uncertainty}
\begin{aligned}
\text{left multiplication}:{T}_l&=\hat{T}\exp_G(\mathcal{L}{(\varepsilon_l)})=\hat{T}\exp_G(\mathcal{L}{(\varepsilon_l)})\hat{T}^{-1}\hat{T}=\exp_G(\mathcal{L}{(Ad_{\hat{T}}(\varepsilon_l))})\hat{T}\\
\text{right multiplication}:{T}_r&=\exp_G(\mathcal{L}{(\varepsilon_r)})\hat{T}
\end{aligned}
\end{equation}

Therefore, the probability distributions for the random variables $T\in SE_K(3)$ can be defined as left-invariant concentrated Gaussian distribution on $SE_K(3)$ and right-invariant concentrated Gaussian distribution on $SE_K(3)$:
\begin{equation}\label{concentrated}
\begin{aligned}
\text{left-invariant}:T\sim \mathcal{N}_L(\hat{T},P),{T}_l&=\hat{T}\exp_G(\mathcal{L}{(\varepsilon_l)}),\varepsilon_l\sim \mathcal{N}(0,P)\\
\text{right-invariant}:T\sim  \mathcal{N}_R(\hat{T},P),{T}_r&=\exp_G(\mathcal{L}{(\varepsilon_r)})\hat{T},\varepsilon_r\sim   \mathcal{N}(0,P)
\end{aligned}
\end{equation}
where $\mathcal{N}(\cdot,\cdot)$ is the classical Gaussian distribution in Euclidean space and $P\in\mathbb{R}^{3(K+1)\times 3(K+1)}$ is a covariance matrix. The invariant property can be verified by $\exp_G(\mathcal{L}{(\varepsilon_r)})=({T}_r \Gamma) (\hat{T}\Gamma)^{-1}={T}_r \hat{T}^{-1}$ and $\exp_G(\mathcal{L}{(\varepsilon_l)})=(\Gamma\hat{T})^{-1}(\Gamma{T}_l ) =\hat{T}^{-1}{T}_l$. The noise-free quantity $\hat{T}$ is viewed as the mean, and the dispersion arises through left multiplication or right multiplication with the matrix exponential of a zero mean Gaussian random variable.
\section{Riemannian Metric, Projection, and dual adjoint}
\begin{definition}
	(Right- and Left- Invariant Riemannian Metric~\cite{bullo2005geometric}). A Riemannian matrix $<\cdot,\cdot>_{X}$ is right-invariant if the differential map of the right translation is an isometry between tangent spaces, that is
	\begin{equation}\label{right_invariant_Riemannian_metric_definition}
	<V,U>_{X}=<T_{X}R_{Y}(V),T_{X}R_{Y}(U)>_{XY}
	\end{equation}
	for all $X,Y\in G$ and $V,U\in T_{X}G$, and is left-invariant if the differential map of the left translation is an isometry between tangent spaces, that is
		\begin{equation}\label{left_invariant_Riemannian_metric_definition}
	<V,U>_{X}=<T_{X}L_{Y}(V),T_{X}L_{Y}(U)>_{YX}
	\end{equation}
	for all $X,Y\in G$ and $V,U\in T_{X}G$.
\end{definition}
\begin{definition}
	(Bi-Invariant Riemannian metric~\cite{bullo2005geometric}). A Riemannian metric is bi-invariant if it is both right-invariant and left-invariant.
\end{definition}

	It is thus uniquely determined by an inner product on the tangent space at the identity $T_eG=\mathfrak{g}$.
	To find a Riemannian metric on the Lie group, an inner product on the Lie algebra is first found.
	An inner product on matrix Lie algebra $\mathfrak{g}$ is denoted as $<\cdot,\cdot>_{\mathfrak{g}}:\mathfrak{g}\times\mathfrak{g}\rightarrow \mathbb{R}$ with associated norm $||\cdot||_{\mathfrak{g}}=\sqrt{<\cdot,\cdot>}$. The standard matrix inner product is given by $<X,Y>=tr(X^TY)$,$\forall X,Y\in\mathbb{R}^{n\times n}$.

For all $X\in SE_{K}(3)$, $A_1, A_2\in \mathfrak{se}_k(3)$, a right-invariant Riemannian metrix $<\cdot,\cdot>_{X}$ on Lie group can be deduced by the inner product on the Lie algebra $\mathfrak{g}$ by the following relationship
\begin{equation}\label{right_invariant_Riemannian_metric}
<A_1X,A_2X>_{X}:=<A_1XX^{-1},A_2XX^{-1}>=<A_1,A_2>,\forall X\in G
\end{equation}

When the Riemannian metric is assumed to be bi-invariant, it is also assumed that the inner product on $\mathfrak{g}$ gives rise to a bi-invariant Riemannian metric by the same relationship~\cite{zlotnik2018lie}.

\begin{definition}
	A Lie group for which $\det(Ad_{g})=1, \forall g\in G$ is called unimodular~\cite{chirikjian2011stochastic}. The Lie group is unimodular means the determinants of the left and right Jacobians are the same since $J_l=Ad_{g}J_r$.
\end{definition}
\begin{remark}
	Since the special orthogonal group $SO(n)$ is compact, it admits a bi-invariant measure, hence $\det(Ad_{R})=\det
	(R)=1$.
\end{remark}
\begin{remark}
	The special Euclidean group $SE(n)$ is unimodular, that is, admits a biinvariant measure. To see that $SE(n)$ is unimodular, consider a left-invariant volume form $\omega$, the volume form is bi-invariant if and only if 
	\begin{equation}\label{bi_invariant_volume_form}
	dL_g\circ dR_{g^{-1}}(\omega_{e})=\omega_{e}
	\end{equation}
	or equivalently $\det(dL_{g}\circ dR_{g^{-1}})=\det(Ad_g)=1$.
\end{remark}
\begin{remark}
	The structure of the adjoint matrix of the Euclidean group $SE(n)$ implies that the derivative of the exponential admits an explict expression.
\end{remark}

For any $g\in SE_K(3)$, we define $|g|_{I}$ as the distance with respect to $I_{K+3}$, which is given by 
\begin{equation}\label{distance_from_unit_element}
|g|_{I}:=||I_{k+3}-g||_{F}=\sqrt{||I_3-R||_{F}^2+\sum_{i=1}^{K}||p_i||^2}=\sqrt{8||R||_{I}^2+\sum_{i=1}^{K}||p_i||^2}
\end{equation}
where $|R|_{I}$ denotes normalized attitude norm on $R\in SO(3)$ with respect to the identity $I_3$~\cite{wang2020geometric}:
\begin{equation}\label{rotation_distance}
|R|_{I}
=\frac{||I_3-R||_{F}}{\sqrt{8}}=\frac{1}{2}\sqrt{tr(I_3-R)}\in[0,1]
\end{equation} 

Let $\mathcal{P}_{\mathfrak{se}_k(3)}:\mathbb{R}^{(K+1)\times (K+1)}\rightarrow \mathfrak{se}_k(3)$ denotes the projection of $A\in \mathbb{R}^{(K+1)\times (K+1)}$ on the Lie algebra $\mathfrak{se}_k(3)$, such that, for $B\in \mathbb{R}^{3\times 3}$, $a_1,a_2,\cdots,a_{2K}\in\mathbb{R}^3$ and $b_1,b_2,\cdots,b_{K^2}\in\mathbb{R}$, one has
\begin{equation}\label{projection_matrix_A_to_lie_algebra}
\begin{aligned}
&\mathcal{P}_{\mathfrak{se}_k(3)}(A)=\mathcal{P}_{\mathfrak{se}_k(3)}\left(
\begin{bmatrix}
B & a_1&a_2&\cdots&a_K\\
a_{K+1}^T&b_1&b_2&\cdots& b_K\\
a_{K+2}^T&b_{K+1}&b_{K+2}&\cdots &b_{2K}\\
\vdots&\vdots&\vdots&\ddots&\vdots\\
a_{2K}^T&b_{K(K-1)+1}&b_{{K(K-1)+}2}&\cdots& b_{K^2}
\end{bmatrix}
\right)\\
=&\begin{bmatrix}
\mathcal{P}_{\mathfrak{so}(3)}(B) &a_1&a_2&\cdots&a_K\\
0^T&0&0&\cdots& 0\\
0^T&0&0&\cdots &0\\
\vdots&\vdots&\vdots&\ddots&\vdots\\
0^T&0&0&\cdots& 0
\end{bmatrix}
\end{aligned}
\end{equation}
where $\mathcal{P}_{\mathfrak{so}_3}:\mathbb{R}^{3\times 3}\rightarrow \mathfrak{so}_3$ is the anti-symmetric projection operator~\cite{zlotnik2018lie,wang2020geometric}.

It follows that, $\forall U\in \mathfrak{se}_k(3)$, $A\in \mathbb{R}^{(K+1)\times (K+1)}$ one has $<U,A>=<U,\mathcal{P}_{\mathfrak{se}_k(3)}(A)>=<\mathcal{P}_{\mathfrak{se}_k(3)}(A),U>=<A,U>$. 

The Adjoint representation of Lie group $G$ on its Lie algebra $\mathfrak{g}$ is defined as the linear operator $Ad_T$:
\begin{equation}\label{Adjoint_representation}
Ad_{T}\cdot \xi:=\mathcal{L}^{-1}\left( T\mathcal{L}(\xi)T^{-1}\right),\forall T\in G,\xi\in\mathbb{R}^p
\end{equation}
where $\mathcal{L}^{-1}$ is the inverse isomorphism of the map $\mathcal{L}$.
It captures the property related to commutation and is equivalent with equation (\ref{adjoint_representation_definition}). 

Specifically, given a rigid body with configuration $T\in SE_K(3)$, the  Adjoint mapping $Ad_T: SE_K(3)\times \mathfrak{se}_k(3)\rightarrow\mathfrak{se}_k(3)$ is given by
\begin{equation}\label{left_action_on_Lie_algebra}
Ad_TU:=TUT^{-1},\forall T\in SE_K(3), U\in\mathfrak{se}_k(3)
\end{equation}
The adjoint mapping allows us to linearly and exactly transforms vectors from the tangent space about one group element to the differential tangent space about another group element.
It is easy to verify that $Ad_{T_1}Ad_{T_2}=Ad_{T_1T_2},\forall T_1,T_2\in SE_K(3)$.

The cotangent space of matrix Lie group $SE_K(3)$ is denoted as $\mathfrak{se}_k^{*}(3)$. For the case $K=1$, the elements of $\mathfrak{se}_k^{*}(3)$ is denoted as $\xi_{\wedge}^*$ and can be identified by a pair $({f},{m})$, where $f\in\mathbb{R}^3$ and $m\in\mathbb{R}^3$ can are the force and moment vector, respectively.
Furthermore, $\xi^*$ is the wrench and its matrix form is given as
\begin{equation}\label{wrench}
\xi_{\wedge}^*=\begin{bmatrix}
f_{\wedge} & m \\
0_{1\times 3} & 0
\end{bmatrix}
\end{equation}


Finally, we consider the dual mapping $ad_{\mathfrak{g}}^{*}$ and its associated Lie group $Ad_{T}^{*}$. 
The metric dual to the adjoint mapping is defined such that~\cite{guigui2021reduced}
\begin{equation}\label{dual_adjoint_algebra}
\forall U,V,W\in \mathfrak{g},<ad_{U}^{*}(V),W>=<V,ad_{U}(W)>=<[U,W],V>
\end{equation}
As the bracket can be computed explicitly in the Lie algebra, so can $ad_{\mathfrak{g}}^{*}$ thanks to the orthogonal basis of $\mathfrak{g}$.

As usual, given a basis for the vector space of the Lie algebra, the matrix representation $ad_{\mathfrak{g}}^{*}$ is simply the transpose of $ad_{\mathfrak{g}}$~\cite{bullo2005geometric}. The dual adjoint mapping $ad_{\mathfrak{g}}^{*}$ is given by
\begin{equation}\label{dual_adjoint}
ad_{\mathfrak{g}}^{*}=(ad_{\mathfrak{g}})^T=\begin{bmatrix}
-\phi_{\wedge}           & -(t_1)_{\wedge}&-(t_2)_{\wedge}&\cdots& -(t_K)_{\wedge}\\
0_{3\times 3} &-\phi_{\wedge}  &0_{3\times 3} &\cdots &0_{3\times 3}\\
0_{3\times 3} &0_{3\times 3} &-\phi_{\wedge}  &\cdots &0_{3\times 3}\\
\vdots&\vdots&\vdots&\ddots&\vdots\\
0_{3\times 3} &0_{3\times 3} &0_{3\times 3}&\cdots &-\phi_{\wedge}  \\
\end{bmatrix}
\end{equation}

The dual Adjoint mapping $Ad_{T}^{*}: SE_K(3)\times \mathfrak{se}_k^*(3)\rightarrow\mathfrak{se}_k^*(3)$ is defined by
\begin{equation}\label{left_action_on_Lie_algebra_dual}
Ad_T^*U:=T^{-1}UT,\forall T\in SE_K(3), U\in\mathfrak{se}_k^*(3)
\end{equation}

Similarly, the matrix representation $Ad_{T}^{*}$ is the simply the transpose of $Ad_{T}$. 
The matrix representation of the dual Adjoint mapping $Ad_{T}^{*}$ is given by
\begin{equation}\label{dual_Adjoint}
Ad_{T}^{*}=(Ad_{T})^T=\begin{bmatrix}
R^T           & -R^T(p_1)_{\wedge}&-R^T(p_2)_{\wedge}&\cdots& -R^T(p_K)_{\wedge}\\
0_{3\times 3} &R^T&0_{3\times 3} &\cdots &0_{3\times 3}\\
0_{3\times 3} &0_{3\times 3} &R^T&\cdots &0_{3\times 3}\\
\vdots&\vdots&\vdots&\ddots&\vdots\\
0_{3\times 3} &0_{3\times 3} &0_{3\times 3}&\cdots &R^T\\
\end{bmatrix}
\end{equation}
%
\section{Vectorial Parameterizations of Matrix Lie Group $SE_K(3)$}
Building on the work of Barfoot~\cite{barfoot2021vectorial}, we show that there also exist several options for mapping vectors onto the matrix Lie group $SE_K(3)$.

Motivated by the general vector pose mapping given by~\cite{barfoot2021vectorial}, we consider the general vector extended pose mapping to be of the same form
\begin{equation}\label{general_vector_extended_post}
T(\xi)=I+a\mathcal{L}(\xi)+b\mathcal{L}(\xi)^2+c\mathcal{L}(\xi)^3
\end{equation}
for some a, b, and c such that the result is an element of the matrix Lie group $SE_K(3)$. 
Owing to the equation (\ref{direct_series_expression}), we can know that any higher-order term of $\mathcal{L}(\xi)$ will be reduced to a lower-order term.

Therefore, we can formulate the explicit expression of $T(\xi)$ by substituting the definition of $\mathcal{L}(\xi)$ in equation (\ref{lie_algebra_se_3_k}) into equation (\ref{general_vector_extended_post}):
\begin{equation}\label{explicit_expression}
\begin{aligned}
T(\xi)&=\begin{bmatrix}
R(\phi) &D(\phi)t_1 &\cdots  &D(\phi)t_K\\
0^T &1&\cdots&0\\
\vdots& \vdots&\ddots&\vdots\\
0^T&0&\cdots&1
\end{bmatrix}\\
&=\begin{bmatrix}
I+(a-\theta^2c)\phi_{\wedge}+b\phi_{\wedge}^2 & (aI+b\phi_{\wedge}+c\phi_{\wedge}^2)t_1&\cdots &(aI+b\phi_{\wedge}+c\phi_{\wedge}^2)t_K\\
 0^T &1&\cdots&0\\
\vdots& \vdots&\ddots&\vdots\\
0^T&0&\cdots&1
\end{bmatrix}
\end{aligned}
\end{equation}
where equation (\ref{cayley_hamilton_rotation}) is used to simplify the top-left entry.

It is obvious that the above result is the direct extension of pose mapping of $SE(3)$. The results of vector mappings for poses can be applied directly here. Therefore, the selection of $a$ and $c$ means the general pose mapping becomes
\begin{equation}\label{extended_pose_mapping}
T(\xi)=I+\mu\mathcal{L}(\xi)+\frac{\nu^2}{2}\mathcal{L}^2+\frac{1}{\theta^2}\left(\mu-\frac{\nu^2}{\varepsilon}\right)(\mathcal{L}(\xi))^3
\end{equation}
 
 When the rotation angle, $\theta$, becomes small, we have the infinitesimal expression
 \begin{equation}\label{order_order_approximation}
 T(\xi)\approx I+\mathcal{L}(\xi)
 \end{equation} 
 This approximation is important and have been used in the perturbations such as equation (\ref{perturbation_kinematic}).
 
 With the definition of \textbf{adjoint} and \textbf{Adjoint}, we can transform the vector parameterization of general pose to its Adjoint in two equivalent ways.
 To verify this commutative property, the series form of the Adjoint mapping can be written as
 \begin{equation}\label{series_form_adjoint_mapping}
 \mathcal{T}(\xi)=I+d\mathfrak{L}(\xi)+e\mathfrak{L}(\xi)^2+f\mathfrak{L}(\xi)^3+g\mathfrak{L}(\xi)^4
 \end{equation}
 for some unknowns, $d$, $e$, $f$, $g$. The higher-order terms can be reduced due to the equation (\ref{direct_series_adjoint_representation}).
 
 Exponential function is one possibility for local parameterization with the advantages that it can be used for any matrix Lie group such that it is applied as standard parameterization. 
 Nevertheless, the usage of the exponential function suffers from one big disadvantage that the inverse of the derivative of the exponential function is an infinite series~\cite{wandelt2019geometric}.
\section{Composition of general poses}
Consider two uncertain general poses $T_{ij}$ and $T_{jk}$ with perturbations $\xi_{ij}$ and $\xi_{jk}$ that are jointly Gaussian in Lie algebra and its covariance matrix is given as
\begin{equation}\label{composition_covariance}
\Sigma=\begin{bmatrix}
\Sigma_{ij} & \Sigma_{ij,jk}\\
\Sigma_{ij,jk}^T & \Sigma_{jk}
\end{bmatrix}
\end{equation}

Denote the composition of the general pose as $T_{ik}$ and its associated mean and covariance as $\left\{\overline{T}_{ik},\Sigma_{ik} \right\}$. 
Under the Lie group multiplication operation, we have
\begin{equation}\label{multiplication_conposition}
{T}_{ik}=T_{ij}T_{jk}
\end{equation}

Following the left uncertainties definition on matrix Lie group, we can get
\begin{equation}\label{perturbation_composition}
\exp_G(\mathcal{L}(\xi_{ik}))\overline{T}_{ik}=\exp_G(\mathcal{L}(\xi_{ij}))\overline{T}_{ij}\exp_G(\mathcal{L}(\xi_{jk}))\overline{T}_{jk}
\end{equation}

Moving all the uncertain factors to the left side by the Adjoint property of the matrix Lie group, we have
\begin{equation}\label{uncertainties_left}
\exp_G(\mathcal{L}(\xi_{ik}))\overline{T}_{ik}=\exp_G(\mathcal{L}(\xi_{ij}))\exp_G(\mathcal{L}(Ad_{\overline{T}_{ij}}\xi_{jk}))\overline{T}_{ij}\overline{T}_{jk}
\end{equation}

If we let
\begin{equation}\label{compostition_extednded_pose}
\overline{T}_{ik}=\overline{T}_{ij}\overline{T}_{jk}
\end{equation}
gives us
\begin{equation}\label{algebra_composition}
\exp_G(\mathcal{L}(\xi_{ik}))=\exp_G(\mathcal{L}(\xi_{ij}))\exp_G(\mathcal{L}(Ad_{\overline{T}_{ij}}\xi_{jk}))
\end{equation}

Letting $\xi_{jk}'=Ad_{\overline{T}_{ij}}\xi_{jk}$, we can use the Baker-Campbell-Housdorff (BCH) formula to show that
\begin{equation}\label{algebra_composition_new}
\begin{aligned}
\xi_{ik}=&\xi_{ij}+\xi_{jk}'+\frac{1}{2}\mathfrak{L}(\xi_{ij})\xi_{jk}'+\frac{1}{12}\mathfrak{L}(\xi_{ij})\mathfrak{L}(\xi_{ij})\xi_{jk}'\\
&+\frac{1}{12}\mathfrak{L}(\xi_{jk}')\mathfrak{L}(\xi_{jk}')\xi_{ij}-\frac{1}{24}\mathfrak{L}(\xi_{jk}')\mathfrak{L}(\xi_{ij})\mathfrak{L}(\xi_{ij})\xi_{jk}'+\cdots
\end{aligned}
\end{equation}

In order to compute the covariance $\Sigma_{ik}$, we have to multiply out up-to fourth order: 
\begin{equation}\label{composition_covariance_sigma}
\begin{aligned}
&\mathbb{E}[\xi_{ik}\xi_{ik}^T]\approx \mathbb{E}[\underbrace{\xi_{ij}\xi_{ij}^T+\xi_{jk}'\xi_{jk}'^T}_{\text{2nd Order Diag. Terms}}+\underbrace{\xi_{ij}\xi_{jk}'^T+\xi_{jk}'\xi_{ij}^T}_{\text{2nd Order Cross Terms}} \\
&+\underbrace{\frac{1}{12}\left({ (\mathfrak{L}(\xi_{ij})\mathfrak{L}(\xi_{ij}))(\xi_{jk}'\xi_{jk}'^T)+(\xi_{jk}'\xi_{jk}'^T)(\mathfrak{L}(\xi_{ij})\mathfrak{L}(\xi_{ij}))^T }\right.}_{\text{4th Order Diagonal Terms}}\\
&\underbrace{\left.{ +(\mathfrak{L}(\xi_{jk}')\mathfrak{L}(\xi_{jk}'))(\xi_{ij}\xi_{ij}^T)+(\xi_{ij}\xi_{ij}^T)(\mathfrak{L}(\xi_{jk}')\mathfrak{L}(\xi_{jk}'))^T   }\right)+\frac{1}{4}\mathfrak{L}(\xi_{ij})(\xi_{jk}'\xi_{jk}'^T)\mathfrak{L}(\xi_{ij})^T}_{\text{4th Order Diagonal Terms}} \\
&+\underbrace{\frac{1}{12} \left({ (\xi_{ij}\xi_{jk}'^T)(\mathfrak{L}(\xi_{ij})^T\mathfrak{L}(\xi_{ij})^T) + (\xi_{jk}'\xi_{ij}^T)(\mathfrak{L}(\xi_{jk}')^T \mathfrak{L}(\xi_{jk}')^T)+(\mathfrak{L}(\xi_{ij})\mathfrak{L}(\xi_{ij}))(\xi_{jk}'\xi_{ij}^T)}\right.}_{\text{4th Order Cross Terms}}\\
&\underbrace{\left.{+(\mathfrak{L}(\xi_{jk}')\mathfrak{L}(\xi_{jk}')(\xi_{ij}\xi_{jk}'^T)}\right)}_{\text{4th Order Cross Terms}}]
\end{aligned} 
\end{equation}
 
If $\xi_{ij}$ and $\xi_{jk}'$ are assumed to be uncorrelated with each other, then we have
\begin{equation}\label{uncorrelated_algebra}
\mathbb{E}[\xi_{ik}]=-\frac{1}{24}\mathfrak{L}(\xi_{jk}')\mathfrak{L}(\xi_{ij})\mathfrak{L}(\xi_{ij})\xi_{jk}'+\mathcal{O}(||\xi_{ik}||^6)
\end{equation}
since everything except the fourth-order term has zero mean~\cite{barfoot2017state}.

As for $\mathbb{E}[\xi_{ik}\xi_{ik}^T]$, when $T_{ij}$ and $T_{jk}$ are independent measurements of consecutive robot motion, the cross terms are zero and the resulting covariance is given as
\begin{equation}\label{fourth_order_terms}
\begin{aligned}
&\mathbb{E}[\xi_{ik}\xi_{ik}^T]\approx \mathbb{E}[\underbrace{\xi_{ij}\xi_{ij}^T+\xi_{jk}'\xi_{jk}'^T}_{\text{2nd Order Diag. Terms}}\\
&+\underbrace{\frac{1}{12}\left({ (\mathfrak{L}(\xi_{ij})\mathfrak{L}(\xi_{ij}))(\xi_{jk}'\xi_{jk}'^T)+(\xi_{jk}'\xi_{jk}'^T)(\mathfrak{L}(\xi_{ij})\mathfrak{L}(\xi_{ij}))^T }\right.}_{\text{4th Order Diagonal Terms}}\\
&\underbrace{\left.{ +(\mathfrak{L}(\xi_{jk}')\mathfrak{L}(\xi_{jk}'))(\xi_{ij}\xi_{ij}^T)+(\xi_{ij}\xi_{ij}^T)(\mathfrak{L}(\xi_{jk}')\mathfrak{L}(\xi_{jk}'))^T   }\right)+\frac{1}{4}\mathfrak{L}(\xi_{ij})(\xi_{jk}'\xi_{jk}'^T)\mathfrak{L}(\xi_{ij})^T}_{\text{4th Order Diagonal Terms}} ]
\end{aligned} 
\end{equation}
where we have omitted showing any terms that have an odd power in either $\xi_{ij}$ and $\xi_{jk}'$  since these will by definition have expectation zero. 

We define two linear operators along the lines of~\cite{barfoot2017state}:
\begin{equation}\label{linear_operator_one}
<<A>>:=-tr(A)I+A
\end{equation}
\begin{equation}\label{linear_operator_two}
<<A,B>>:=<<A>><<B>>+<<BA>>
\end{equation}
where $A,B\in \mathbb{R}^{n\times n}$. These provide the following useful identity,
\begin{equation}\label{identity_phi_omega}
-\phi_{\wedge} D \omega_{\wedge}=<<\omega\phi^T,D>>
\end{equation}
where $\phi,\omega\in\mathbb{R}^3$ and $D\in\mathbb{R}^{3\times 3}$.

Now we compute the equation (\ref{fourth_order_terms}) term by term.
\begin{equation}\label{xi_1}
\mathbb{E}[\xi_{ij}\xi_{ij}^T]=\Sigma_{ij}=\begin{bmatrix}
\Sigma_{ij,\phi\phi} & \Sigma_{ij,\phi t_1} & \Sigma_{ij,\phi t_2}&\cdots & \Sigma_{ij,\phi t_K}\\
\Sigma_{ij,t_1\phi}  & \Sigma_{ij,t_1t_1} & \Sigma_{ij,t_1 t_2} &\cdots & \Sigma_{ij,t_1 t_K}\\
\Sigma_{ij,t_2\phi}   &\Sigma_{ij,t_2 t_1} &\Sigma_{ij,t_2t_2} & \cdots & \Sigma_{ij,t_2 t_K}\\
\vdots &\vdots &\vdots &\ddots & \vdots\\
\Sigma_{ij,t_K\phi}   &\Sigma_{ij,t_Kt_1} &\Sigma_{ij,t_Kt_2}&\cdots& \Sigma_{ij,t_Kt_K}
\end{bmatrix}
\end{equation}
\begin{equation}\label{xi_2_prime}
\mathbb{E}[\xi_{jk}'\xi_{jk}'^T]=\Sigma_{jk}'=\begin{bmatrix}
\Sigma_{jk,\phi\phi}' & \Sigma_{jk,\phi t_1}' & \Sigma_{jk,\phi t_2}'&\cdots & \Sigma_{jk,\phi t_K}'\\
\Sigma_{jk,t_1\phi}'  & \Sigma_{jk,t_1t_1}' & \Sigma_{jk,t_1 t_2}' &\cdots & \Sigma_{jk,t_1 t_K}'\\
\Sigma_{jk,t_2\phi}'   &\Sigma_{jk,t_2 t_1}' &\Sigma_{jk,t_2t_2}' & \cdots & \Sigma_{jk,t_2 t_K}'\\
\vdots &\vdots &\vdots &\ddots & \vdots\\
\Sigma_{jk,t_K\phi}'   &\Sigma_{jk,t_Kt_1}' &\Sigma_{jk,t_Kt_2}'&\cdots& \Sigma_{jk,t_Kt_K}'
\end{bmatrix}=Ad_{\overline{T}_{ij}}\Sigma_{jk}Ad_{\overline{T}_{ij}}^T
\end{equation}

We first compute
\begin{equation}\label{xi__xi_mathfrak}
\mathfrak{L}(\xi_{ij})\mathfrak{L}(\xi_{ij})=\begin{bmatrix}
\phi_{\wedge}\phi_{\wedge} & 0 & 0&\cdots & 0\\
(t_1)_{\wedge}\phi_{\wedge}+\phi_{\wedge}(t_1)_{\wedge} & \phi_{\wedge}\phi_{\wedge} & 0 &\cdots & 0\\
(t_2)_{\wedge}\phi_{\wedge}+\phi_{\wedge}(t_2)_{\wedge} &0 & \phi_{\wedge}\phi_{\wedge} & \cdots & 0\\
\vdots &\vdots &\vdots &\ddots & \vdots\\
(t_K)_{\wedge}\phi_{\wedge}+\phi_{\wedge}(t_K)_{\wedge} &0 &0&\cdots& \phi_{\wedge}\phi_{\wedge}
\end{bmatrix}
\end{equation}

Making use of the property 
\begin{equation}\label{phi_phi_property}
\phi_{\wedge}\phi_{\wedge}=-(\phi^T\phi)I_3+\phi\phi^T
\end{equation}

and taking the expectation of equation (\ref{xi__xi_mathfrak}), we obtain
\begin{equation}\label{xi__xi_mathfrak_expectation}
\begin{aligned}
&\mathbb{E}[\mathfrak{L}(\xi_{ij})\mathfrak{L}(\xi_{ij})]=\mathcal{A}_1\\
=&\begin{bmatrix}
<<\Sigma_{ij,\phi\phi}>> & 0 & 0&\cdots & 0\\
<<\Sigma_{ij,t_1\phi} +\Sigma_{ij,\phi t_1}>> & <<\Sigma_{ij,\phi\phi}>> & 0 &\cdots & 0\\
<<\Sigma_{ij,t_2\phi} +\Sigma_{ij,\phi t_2}>>  &0 &<<\Sigma_{ij,\phi\phi}>> & \cdots & 0\\
\vdots &\vdots &\vdots &\ddots & \vdots\\
<<\Sigma_{ij,t_K\phi} +\Sigma_{ij,\phi t_K}>>  &0 &0&\cdots& <<\Sigma_{ij,\phi\phi}>>
\end{bmatrix}\\
=&\begin{bmatrix}
<<\Sigma_{ij,\phi\phi}>> & 0 & 0&\cdots & 0\\
<<\Sigma_{ij,t_1\phi} +\Sigma_{ij,t_1\phi}^T>> & <<\Sigma_{ij,\phi\phi}>> & 0 &\cdots & 0\\
<<\Sigma_{ij,t_2\phi} +\Sigma_{ij,t_2\phi}^T>>  &0 &<<\Sigma_{ij,\phi\phi}>> & \cdots & 0\\
\vdots &\vdots &\vdots &\ddots & \vdots\\
<<\Sigma_{ij,t_K\phi} +\Sigma_{ij,t_K\phi}^T>>  &0 &0&\cdots& <<\Sigma_{ij,\phi\phi}>>
\end{bmatrix}
\end{aligned}
\end{equation}

Similarly, we can obtain
\begin{equation}\label{xi__xi_prime_mathfrak_expectation}
\begin{aligned}
&\mathbb{E}[\mathfrak{L}(\xi_{jk}')\mathfrak{L}(\xi_{jk}')]=\mathcal{A}_2'\\
=&\begin{bmatrix}
<<\Sigma_{jk,\phi\phi}'>> & 0 & 0&\cdots & 0\\
<<\Sigma_{jk,t_1\phi}' +\Sigma_{jk,\phi t_1}'>> & <<\Sigma_{jk,\phi\phi}'>> & 0 &\cdots & 0\\
<<\Sigma_{jk,t_2\phi}'+\Sigma_{jk,\phi t_2}'>>  &0 &<<\Sigma_{jk,\phi\phi}'>> & \cdots & 0\\
\vdots &\vdots &\vdots &\ddots & \vdots\\
<<\Sigma_{jk,t_K\phi}' +\Sigma_{jk,\phi t_K}'>>  &0 &0&\cdots& <<\Sigma_{jk,\phi\phi}'>>
\end{bmatrix}\\
=&\begin{bmatrix}
<<\Sigma_{jk,\phi\phi}'>> & 0 & 0&\cdots & 0\\
<<\Sigma_{jk,t_1\phi}' +\Sigma_{jk,t_1\phi}'^T>> & <<\Sigma_{jk,\phi\phi}'>> & 0 &\cdots & 0\\
<<\Sigma_{jk,t_2\phi}'+\Sigma_{jk,t_2\phi}'^T>>  &0 &<<\Sigma_{jk,\phi\phi}'>> & \cdots & 0\\
\vdots &\vdots &\vdots &\ddots & \vdots\\
<<\Sigma_{jk,t_K\phi}' +\Sigma_{jk,t_K\phi}'^T>>  &0 &0&\cdots& <<\Sigma_{jk,\phi\phi}'>>
\end{bmatrix}
\end{aligned}
\end{equation}

As the noises are not correlated, we can obtain 
\begin{equation}\label{fourth_order_term}
\begin{aligned}
&\mathbb{E}[
(\mathfrak{L}(\xi_{ij})\mathfrak{L}(\xi_{ij}))(\xi_{jk}'\xi_{jk}'^T)+(\xi_{jk}'\xi_{jk}'^T)(\mathfrak{L}(\xi_{ij})\mathfrak{L}(\xi_{ij}))^T +(\mathfrak{L}(\xi_{jk}')\mathfrak{L}(\xi_{jk}'))(\xi_{ij}\xi_{ij}^T)+(\xi_{ij}\xi_{ij}^T)(\mathfrak{L}(\xi_{jk}')\mathfrak{L}(\xi_{jk}'))^T ]\\
=&\mathcal{A}_1\Sigma_{jk}'+\Sigma_{jk}'\mathcal{A}_1^T+\mathcal{A}_2'\Sigma_{ij}+\Sigma_{ij}\mathcal{A}_2'^T
\end{aligned}
\end{equation}

It now remains to compute
\begin{equation}\label{B_complex_definition}
\begin{aligned}
\mathbb{E}[\mathfrak{L}(\xi_{ij})(\xi_{jk}'\xi_{jk}'^T)\mathfrak{L}(\xi_{ij})^T]=\mathcal{B}=\begin{bmatrix}
B_{\phi\phi} & B_{\phi t_1}& B_{\phi t_2}& \cdots & B_{\phi t_K}\\
B_{t_1\phi} & B_{t_1 t_1}& B_{t_1 t_2}& \cdots & B_{t_1 t_K}\\
B_{t_2\phi} & B_{t_2 t_1}& B_{t_2 t_2}& \cdots & B_{t_2 t_K}\\
\vdots  & \vdots & \vdots &\ddots & \vdots\\
B_{t_K\phi} & B_{t_K t_1}& B_{t_K t_2}& \cdots & B_{t_K t_K}\\
\end{bmatrix}
\end{aligned}
\end{equation}

As variables are not correlated, we have
\begin{equation}\label{B_complex_calculation}
\mathcal{B}=\mathbb{E}[\mathfrak{L}(\xi_{ij})(\xi_{jk}'\xi_{jk}'^T)\mathfrak{L}(\xi_{ij})^T]=\mathbb{E}[\mathfrak{L}(\xi_{ij})\mathbb{E}[(\xi_{jk}'\xi_{jk}'^T)]\mathfrak{L}(\xi_{ij})^T]=\mathbb{E}[\mathfrak{L}(\xi_{ij})\Sigma_{jk}'\mathfrak{L}(\xi_{ij})^T]
\end{equation}

Developing the above expression and taking the expectation by using the relation (\ref{linear_operator_two}), we obtain
\begin{equation}\label{B_phi_calculation}
B_{\phi\phi}=<<\Sigma_{ij,\phi\phi},\Sigma_{jk,\phi\phi}'>>
\end{equation}
\begin{equation}\label{B_t_K_phi_calculation}
\begin{aligned}
B_{t_m \phi}=<<\Sigma_{ij,\phi t_m},\Sigma_{jk,\phi\phi}'>>+<<\Sigma_{ij,\phi\phi},\Sigma_{jk,t_m\phi }'>>,t_m=t_1,t_2,\cdots,t_K
\end{aligned}
\end{equation}
\begin{equation}\label{B_phi_t_K_calculation}
\begin{aligned}
B_{\phi t_m}=B_{t_m \phi}^T=<<\Sigma_{ij, t_m \phi},\Sigma_{jk,\phi\phi}'>>+<<\Sigma_{ij,\phi\phi},\Sigma_{jk,\phi t_m }'>>,t_m=t_1,t_2,\cdots,t_K
\end{aligned}
\end{equation}
\begin{equation}\label{B_t_K_t_K_calculation}
\begin{aligned}
B_{t_m,t_m}
=&<<\Sigma_{ij,t_mt_m},\Sigma_{jk,\phi\phi}'>>+<<\Sigma_{ij,t_m\phi},\Sigma_{jk,t_m\phi}'>>\\
&+<<\Sigma_{ij,\phi t_m  },\Sigma_{jk,\phi t_m}'>>+<<\Sigma_{ij,\phi\phi},\Sigma_{jk,t_m t_m}'>>
,t_m=t_1,t_2,\cdots,t_K
\end{aligned}
\end{equation}
\begin{equation}\label{t_m_left_down_t_n}
\begin{aligned}
B_{t_m,t_n}=&<<\Sigma_{ij,t_{n}t_{m}},\Sigma_{jk,\phi\phi}'>>+<<\Sigma_{ij,t_{n}\phi},\Sigma_{jk,t_{m}\phi}'>>\\
&+<<\Sigma_{ij,\phi t_{m}},\Sigma_{jk,\phi t_{n}}'>>+<<\Sigma_{ij,\phi\phi},\Sigma_{jk,t_{m}t_{n}}'>>,\\
&t_m=t_2,t_3,\cdots,t_K,t_n=t_1,t_2,\cdots,t_{K-1}, t_m>t_n
\end{aligned}
\end{equation}
\begin{equation}\label{t_m__right_updown_t_n}
\begin{aligned}
B_{t_n,t_m}=B_{t_m,t_n}^T=&<<\Sigma_{ij,t_{m}t_{n}},\Sigma_{jk,\phi\phi}'>>+<<\Sigma_{ij,t_{m}\phi},\Sigma_{jk,t_{n}\phi}'>>\\
&+<<\Sigma_{ij,\phi t_{n}},\Sigma_{jk,\phi t_{m}}'>>+<<\Sigma_{ij,\phi\phi},\Sigma_{jk,t_{n}t_{m}}'>>,\\
&t_n=t_1,t_2,\cdots,t_{K-1},t_m=t_2,t_3,\cdots,t_K, t_n<t_m
\end{aligned}
\end{equation}

We finally get the formula for equation (\ref{fourth_order_terms}):
\begin{equation}\label{finally_expectation}
\mathbb{E}[\xi_{ik}\xi_{ik}^T]\approx \underbrace{\Sigma_{ij}+\Sigma_{jk}'}_{\Sigma_{2nd}}+ \underbrace{\frac{1}{12}\left( \mathcal{A}_1\Sigma_{jk}'+\Sigma_{jk}'\mathcal{A}_1^T+\mathcal{A}_2'\Sigma_{ij}+\Sigma_{ij}\mathcal{A}_2'^T\right)+\frac{1}{4}\mathcal{B}}_{\text{additional fourth-order terms}}
\end{equation} 

In summary, to compound two general poses, the mean and the covariance can be propagates by equation (\ref{compostition_extednded_pose}) and equation (\ref{finally_expectation}), respectively.

However, if the general poses $T_{ij}$ and $T_{jk}$ are correlated, as is often the case when they are derived from the solution of a maximum likelihood estimate problem such as Pose Graph SLAM or in the case of wheel slip, the cross terms must be included and evaluation of the fourth order terms becomes more difficult~\cite{mangelson2019characterizing}. The general pose composition operation may under-approximate the true distribution and lead to inconsistency problem if the cross term is ignored.

The first order covariance calculation can be obtained by evaluating 
\begin{equation}\label{first_order_covariance}
\mathbb{E}[\xi_{ik}\xi_{ik}^T]\approx \mathbb{E}[\xi_{ij}\xi_{ij}^T+\xi_{jk}'\xi_{jk}'^T+\xi_{ij}\xi_{jk}'^T+\xi_{jk}'\xi_{ij}^T]
\end{equation}
which means
\begin{equation}\label{one_order_covariance_aproximation}
\Sigma_{ik}\approx \Sigma_{ij}+Ad_{\overline{T}_{ij}}\Sigma_{jk}Ad_{\overline{T}_{ij}}^T+\Sigma_{ij,jk}Ad_{\overline{T}_{ij}}^T+Ad_{\overline{T}_{ij}}\Sigma_{ij,jk}^T
\end{equation}

Thus, for the first-order general pose composition on the matrix Lie algebra can be performed for correlated poses using equation (\ref{compostition_extednded_pose}) for mean propagation and equation (\ref{one_order_covariance_aproximation}) for covariance propagation.
\section{Monotonicity of Uncertainty Propagation of Matrix Lie group $SE_K(3)$}
Considering only the second-order terms in equation (\ref{finally_expectation}), we can obtain that
\begin{equation}\label{second_order_term}
\Sigma_{ik}=\mathbb{E}[\xi_{ik}\xi_{ik}^T]\approx \Sigma_{ij}+\Sigma_{jk}'=\underbrace{\Sigma_{ij}}_{\text{initial uncertainty}}+\underbrace{Ad_{\overline{T}_{ij}}\Sigma_{jk}Ad_{\overline{T}_{ij}}^T}_{\text{motion's uncertainty}}
\end{equation}
It is obvious that the uncertainty on matrix Lie group $SE_K(3)$ consists of the initial uncertainty and motion's uncertainty.
Furthermore, the initial uncertainty keeps still and is independent of the system's trajectory even the motion's uncertainty varies as the vehicle travels~\cite{kim2017uncertainty}.

The motion's uncertainty is related with the Adjoint operator of the transformation $\overline{T}_{ij}$. Since the uncertainty matrix $\Sigma_{jk}$ is positive definite, the motion's uncertainty $Ad_{\overline{T}_{ij}}\Sigma_{jk}Ad_{\overline{T}_{ij}}^T$ can be shown to be positive definite. The proof is given as following
\begin{equation}\label{positive_definite_proof}
x^TAd_{\overline{T}_{ij}}\Sigma_{jk}Ad_{\overline{T}_{ij}}^Tx=(Ad_{\overline{T}_{ij}}^Tx)^T\Sigma_{jk}(Ad_{\overline{T}_{ij}}^Tx)>0,\forall x\in \mathbb{R}^{3(K+1)}
\end{equation} 

Similar to the Theorem 1 in~\cite{kim2017uncertainty}, it can be proven that the monotocinity of uncertainty on matrix Lie group $SE_K(3)$ holds for all the optimality criteria (the A-opt(trace of the covariance matrix, or sum of its eigenvalue), D-opt(determinant of the covariance matrix, or product of ite eigenvalue), and E-opt(largest eigenvalue)). 
\begin{theorem}
	~\cite{kim2017uncertainty}Every eigenvalue of the uncertainty defined on matrix Lie group $SE_K(3)$ as in equation (\ref{second_order_term}) increases, that means eigenvalues of uncertainty matrix $\Sigma_{ik}$ are always greater than the corresponding eigenvalues of uncertainty matrix $\Sigma_{ij}$.
\end{theorem}
\begin{proof}
	Assume the matrices $M$, $H$, and $P$ are Hermitian matrices and have relationship as $M=H+P$. The eigenvalues of the three matrices are given as
	\begin{equation}\label{eigenvalues_M}
	\lambda_m(M)\leq \cdots \leq \lambda_j(M)\leq \cdots \leq \lambda_1(M)
	\end{equation}
	\begin{equation}\label{eigenvalues_H}
	\lambda_m(H)\leq \cdots \leq \lambda_j(H)\leq \cdots \leq \lambda_1(H)
	\end{equation}
	\begin{equation}\label{eigenvalues_P}
		\lambda_m(P)\leq \cdots \leq \lambda_j(P)\leq \cdots \leq \lambda_1(P)
	\end{equation}
	According to the Weyl's inequality\footnote{\url{https://en.wikipedia.org/wiki/Weyl\%27s_inequality}}, the following inequality holds for all $j$:
	\begin{equation}\label{MPH}
	\lambda_j(H)+\lambda_m(P)\leq \lambda_j(M)\leq \lambda_j(H)+\lambda_1(P),(1\leq j\leq m)
	\end{equation}
	In particular, if $P$ is positive definite and $\lambda_m(P)>0$, then
	\begin{equation}\label{MH_eigenvalue}
	\lambda_j(M)>\lambda_j(H),\forall j=1,\cdots, m
	\end{equation}
	Next, the matrix $M$, $H$, and $P$ are considered as uncertainty matrices $\Sigma_{ij}$, $\Sigma_{ik}$, and $Ad_{\overline{T}_{ij}}\Sigma_{jk}Ad_{\overline{T}_{ij}}^T$ in equation (\ref{second_order_term}), respectively. According to equation (\ref{MH_eigenvalue}) we have
	\begin{equation}\label{final_result}
	\lambda_j(\Sigma_{ik})>\lambda_j(\Sigma_{ij})
	\end{equation}
	In other words, every eigenvalue of $\Sigma_{ij}$ is monotonically increasing of the uncertainty propagation on matrix Lie group $SE_K(3)$.
	\end{proof}
The above theorem shows that keeping monotonicity is matter of correctly modeling errors and propagating uncertainty~\cite{kim2017uncertainty}.

Nest, we study the monotonicity of the uncertainty from the perspective of R\'enyi entropy.
The R\'enyi entropy of the multivariate normal distribution can be calculated by
\begin{equation}\label{Renyi_entropy}
H_{\alpha}=\frac{1}{2}\log(\det\Sigma)+\frac{N}{2}\log(2\pi\alpha^{\frac{1}{\alpha-1}}), \alpha\in[0,1)\cup(1,\infty)
\end{equation} 
where $\alpha$ is a free parameter, $N$ is the dimension of the state, $\Sigma$ is the uncertainty matrix.
As $\alpha\rightarrow 1$, the Shannon entropy of the multivariate normal distribution can be obtained.

It can be shown that the monotonicity of the R\'enyi entropy is equivalent to the monotonicity of the D-opt optimality criteria because
\begin{equation}\label{renyi_entropy}
H_{\alpha}(\Sigma_{ik})-H_{\alpha}(\Sigma_{ij})=\frac{1}{2}\log\left(\frac{\det(\Sigma_{ik}) }{\det(\Sigma_{ij}) }  \right)
\end{equation}

Applying the Minkowski's inequality to equation (\ref{second_order_term}), we have
\begin{equation}\label{minkowski}
\det(\Sigma_{ik})>\det(\Sigma_{ij})
\end{equation}

In the end, the monotonicity of the uncertainty propagation for the R\'enyi entropy is maintained:
\begin{equation}\label{monotonicity}
H_{\alpha}(\Sigma_{ik})-H_{\alpha}(\Sigma_{ij})=\frac{1}{2}\log\left(\frac{\det(\Sigma_{ik}) }{\det(\Sigma_{ij}) }  \right)>0
\end{equation}

\section{Applications of $SE_K(3)$}
The application about this group mainly includes three aspects: the 3-dimensional inertial navigation, the IMU preintegration and the simultaneous localization and mapping problem. All applications require that the system dynamics are group affine, becasue group affine systems have log-linear property of the error propagation.
The invariant EKF framework can improve convergence and accuracy while potentially eliminating the
need for re-linearization.

Most of the applications are related to inertial navigation and the error states are easy to be modeled so as to satisfy the group affine property. The measurement equations are determined by the definition form of the error state. When the error state is left invariant by the left group action, the measurement should be also left invariant, this is the world-centric estimator formulation and is suitable for sensors such as GNSS, 5G, etc. When the error state is right invariant by the right group action, the measurement should be also right invariant, this is body-centric formulation and is suitable for sensors such as camera, Lidar, odometry, etc.

\section{Conclusions}
In this paper, the geometry and kinematics of the matrix Lie group $SE_K(3)$ is given as an extension of the matrix Lie group $SE(3)$. The application of this group is analyzed in detail. Finally, we share the simulation code about this group. More applications will be given in future.
\vspace{2ex}

\noindent
{\bf\normalsize Acknowledgement}\newline
{This research was supported by a grant from the National Key Research and Development Program of China (2018YFB1305001). 
We express thanks to GNSS Center, Wuhan University.} \vspace{2ex}

\bibliographystyle{IEEEtran}
\bibliography{ref.bib}

\end{document}